\documentclass[ijoc,sglanonrev]{informs4}
\usepackage{eqndefns-left} 
\RequirePackage{tgtermes}
\RequirePackage{newtxtext}
\RequirePackage{newtxmath}
\RequirePackage{bm}
\RequirePackage{endnotes}

\OneAndAHalfSpacedXII 

\usepackage{bm}
\usepackage{booktabs}
\usepackage{colortbl}
\usepackage{multirow}
\usepackage{multicol}
\usepackage{siunitx}
\usepackage{braket}
\usepackage{algorithm}
\usepackage{algorithmicx}
\usepackage{algpseudocode}
\usepackage{mathtools}
\usepackage{placeins}
\usepackage{tcolorbox}
\usepackage{caption}
\usepackage{subcaption}
\usepackage{amsfonts, amsmath, amssymb}
\usepackage[numbers,square]{natbib}
\usepackage{pifont}
\newcommand{\cmark}{\ding{51}}
\newcommand{\xmark}{\ding{55}}

\usepackage[hidelinks]{hyperref}
\hypersetup{
    colorlinks=true,
    linkcolor=black,
    filecolor=black,      
    urlcolor=black,
    citecolor=black,
}

\newcommand{\convhull}[1]{\mathrm{Conv} \left( #1 \right)}
\newcommand{\indexes}[1]{[#1]}
\newcommand{\rank}[1]{\mathrm{Rank} \left( #1 \right)}
\newcommand{\reals}[1]{\mathbb{R}^{#1}}
\newcommand{\trace}[1]{\mathrm{tr} \left( #1 \right)}
\newcommand{\projs}[2]{\mathcal{Y}^{#2}_{#1}}

\usepackage{natbib}
 \bibpunct[, ]{(}{)}{,}{a}{}{,}%
 \def\bibfont{\small}%
 \def\BIBand{and}%

\EquationsNumberedThrough    

\TheoremsNumberedThrough     
\ECRepeatTheorems  %

\MANUSCRIPTNO{}

\begin{document}


\RUNAUTHOR{Bertsimas et al.}

\RUNTITLE{Optimal Low-Rank Matrix Completion}

\TITLE{Disjunctive Branch-and-Bound for Certifiably Optimal Low-Rank Matrix Completion}
\ARTICLEAUTHORS{%
    \AUTHOR{Dimitris Bertsimas}
    \AFF{
        Sloan School of Management and Operations Research Center, Massachusetts Institute of Technology, Cambridge, MA, USA. \\
        ORCID: 0000-0002-1985-1003 \\
        \EMAIL{dbertsim@mit.edu}
    }
    \AUTHOR{Ryan Cory-Wright}
    \AFF{
        Department of Analytics, Marketing and Operations, Imperial Business School, London, UK.\\
        ORCID: 0000-0002-4485-0619 \\
        \EMAIL{r.cory-wright@imperial.ac.uk}
    }
    \AUTHOR{Sean Lo}
    \AFF{
        Operations Research Center, Massachusetts Institute of Technology, Cambridge, MA, USA. \\ 
        ORCID: 0000-0001-8456-6471 \\
        \EMAIL{seanlo@mit.edu}
    }
    \AUTHOR{Jean Pauphilet}
    \AFF{
        Management Science \& Operations, London Business School, London, UK. \\ \
        ORCID: 0000-0001-6352-0984 \\
        \EMAIL{jpauphilet@london.edu}
    }
} 

\ABSTRACT{%
Low-rank matrix completion consists of computing a matrix of minimal complexity that recovers a given set of observations as accurately as possible. Unfortunately, existing methods for matrix completion are heuristics that, while highly scalable and often identifying high-quality solutions, do not {\color{black} provide an instance-wise certificate of} optimality. We reexamine matrix completion with an optimality-oriented eye. We reformulate low-rank matrix completion problems as convex problems over the non-convex set of projection matrices and implement a disjunctive branch-and-bound scheme that solves them to certifiable optimality. Further, we derive a novel and often near-exact class of convex relaxations by decomposing a low-rank matrix as a sum of rank-one matrices and incentivizing that two-by-two minors in each rank-one matrix have determinant zero. In numerical experiments, our new convex relaxations decrease the optimality gap by two orders of magnitude compared to existing attempts, and our disjunctive branch-and-bound scheme solves $n \times m$ rank-$k$ matrix completion problems to certifiable optimality or near optimality in hours for $\max \{m, n\} \leq 2500$ and $k \leq 5$. Moreover, this {\color{black}reduction} in the training error translates into an average $2\%$--$50\%$ {\color{black}reduction} in the test set error {\color{black}compared with alternating minimization-based methods}.
}%





\maketitle


\section{Introduction} 
This paper proposes a branch-and-bound scheme for solving low-rank matrix completion to provable optimality. Given observations $A_{i,j} : (i,j) \in \mathcal{I} \subseteq [n] \times [m]$ from a matrix $\bm{A} \in \reals{n \times m}$, we seek a low-rank matrix $\bm{X} \in \reals{n \times m}$ which approximates the observed entries of $\bm{A}$. This admits the formulation:
\begin{equation}
\begin{aligned}
    \label{eq:rco_matrix_completion}
    \min_{\bm{X} \in \reals{n \times m}} \quad
    \frac{1}{2 \gamma} \Vert \bm{X} \Vert^2_F+\frac{1}{2} \sum_{(i,j) \in \mathcal{I}} \left( X_{i,j} - A_{i,j} \right)^2
    \quad &
    \text{s.t.} 
    \quad 
    \rank{\bm{X}} \leq k,
\end{aligned}
\end{equation}
where the hyperparameter $k$ bounds the rank of $\bm{X}$ and $\gamma$ controls {\color{black} the degree of regularization}. %
{\color{black} The objective in \eqref{eq:rco_matrix_completion} involves a squared Frobenius norm penalty as done in \citet{keshavan2010regularization,sun2012calibrated}. 
In the literature, such a regularization has been justified to improve out-of-sample prediction {\color{black}and} avoid overfitting in high-noise regimes \citep{keshavan2010regularization} or to make the objective strongly convex in $\bm{X}$ \citep{sun2012calibrated}.
We provide a more detailed discussion on regularization for matrix completion problems in Section \ref{ssec:regularization}.}

Problem \eqref{eq:rco_matrix_completion} has received a great deal of attention since 
the Netflix Competition \citep{bell2007lessons}, and heuristic methods now find high-quality solutions for very large-scale instances. Unfortunately, low-rank problems cannot be modeled as mixed-integer conic optimization problems \citep[see][]{lubin2022mixed} and thus are out of scope for traditional combinatorial optimization algorithms. 
For this reason, developing a certifiably optimal method for low-rank matrix completion would be theoretically meaningful. 
Moreover, from a practical perspective, \citet{MPCO_2021} have shown that 
solving \eqref{eq:rco_matrix_completion} via a provably optimal method gives a 
smaller out-of-sample mean squared error (MSE) than the MSE obtained via a heuristic. However, to the best of our knowledge, no method solves Problem \eqref{eq:rco_matrix_completion} to provable optimality for $n,m$ beyond $50$ or $k$ beyond $1$ \citep{naldi2016solving, MPCO_2021}.

In this paper, we design a custom branch-and-bound method that solves Problem \eqref{eq:rco_matrix_completion} to provable optimality using {\color{black} eigenvector disjunctions} as branching regions. Numerically, we {\color{black} assess the scalability of our approach for instances with $\max \{m, n\}$ up to $2500$ and $k$ up to $5$}. We also show numerically that this approach obtains more accurate solutions than state-of-the-art methods like alternating minimization implementations of the method of \citet{burer2003nonlinear}, {\color{black}by up to $50\%$ for some small-scale instances with many local optima}.

\subsection{Literature Review}\label{ssec:lit.rev}
We propose a branch-and-bound algorithm that solves Problem \eqref{eq:rco_matrix_completion} to provable optimality at scale. To develop this algorithm, we require three ingredients: a strategy for recursively partitioning the solution space; a technique for generating high-quality convex relaxations on each partition; and a local optimization strategy for quickly obtaining good solutions in a given partition. To put our contribution into context, we now review all three aspects of the relevant literature and refer to \cite{udell2016generalized, MPCO_2021} for overviews of low-rank optimization.

\paragraph{Branching} A variety of spatial branch-and-bound schemes have been proposed for generic non-convex quadratically constrained quadratic optimization problems such as \eqref{eq:rco_matrix_completion}, since the work of \citet{mccormick1976computability}. 
Unfortunately, these solvers currently cannot obtain high-quality solutions for problems with more than fifty variables \citep[see][for a benchmark and references]{kronqvist2019review}.  
Recently, \citet{MPCO_2021} reformulated \eqref{eq:rco_matrix_completion} via projection matrices and solved problems with $50 \times 50$ matrices and $k=1$ using off-the-shelf solvers that use McCormick branching regions. 
Problem-specific branching strategies have been investigated for other non-convex QCQPs, {\color{black} such as two-trust-region problems \citep{anstreicher2022solving} or optimal power flow problems \citep{kocuk2018matrix}. Our branching strategy, like that of \citet{anstreicher2022solving}, is based on the idea of computing eigenvectors for non-convex semidefinite constraints that are violated at a relaxed solution, as proposed by \cite{saxena2010convex}. However, while \citet{saxena2010convex} derive cutting planes, we take a different approach and use eigenvectors to develop a branching scheme.}

\paragraph{Relaxations} 
A number of works have proposed solving Problem \eqref{eq:rco_matrix_completion} via their convex relaxations, originating with \citet{fazel2002matrix}, who proposed replacing the rank minimization objective 
with a trace term 
for positive semidefinite factor analysis and matrix completion problems \citep[see also][for an extension to the asymmetric case]{candes2009exact}. 
More recently, \citet{bertsimas2021new} proposed a general procedure for obtaining strong bounds to low-rank problems 
\citep[see also][for related attempts]{kim2021convexification, li2023rank}. Namely, they combined the projection matrix reformulation of \citet{MPCO_2021} with a matrix analog of perspective functions \citep[c.f.][]{ ebadian2011perspectives} 
and obtained a new class of convex relaxations for low-rank problems. 

\paragraph{High-quality heuristics} 
The most popular approach {\color{black}for finding} good solutions to Problem \eqref{eq:rco_matrix_completion} is to enforce the rank constraint via the decomposition $\bm{X}=\bm{U}\bm{V}$, where the matrices $\bm{U}$ and $\bm{V}$ have $k$ columns {\color{black}and $k$ rows respectively}, and to iteratively optimize with respect to $\bm{U}$ and $\bm{V}$ \citep{burer2003nonlinear}. However, the Burer-Monteiro approach is only guaranteed to converge to a local solution and several works have investigated conditions under which convergence to a global optimum is guaranteed for matrix completion \citep{jain2013low,hardt2014understanding,ge2016matrix,ma2023optimization} or generic semidefinite optimization problems \citep{boumal2016non,bhojanapalli2016global,cifuentes2019burer}.
{\color{black} These guarantees require assumptions about the data-generating process that may be hard to verify on a particular instance \citep{bandeira2013certifying} and hold only probabilistically (with high probability). In contrast, a branch-and-bound algorithm provides, for any instance, a feasible solution with an instance-wise certificate of near-optimality (i.e., an optimality gap). 
}



\subsection{Contributions}
We 
develop a spatial branch-and-bound scheme which solves medium-sized instances of Problem~\eqref{eq:rco_matrix_completion} to certifiable (near) optimality. 

The key contributions of the paper are threefold. First, we derive {\color{black} an eigenvector-based branching scheme} that recursively partitions the feasible region and strengthens the matrix perspective relaxations of Problem \eqref{eq:rco_matrix_completion} more effectively than via McCormick relaxations {\color{black} (see Proposition~\ref{prop:mccormickisbad_k})}. 
{\color{black} This approach bears similarity to the eigenvector disjunctive cuts proposed in \citet{saxena2010convex}; however, instead of a cutting-plane approach, we use our eigenvector disjunctions to construct a spatial branching scheme.
}
{\color{black} 
Second, we implement this branching strategy in a complete branch-and-bound algorithm. 
}
Third, by combining an old characterization of rank via determinant minors, which is new to the matrix completion literature, with the Shor relaxation, we derive new and tighter convex relaxations for matrix completion problems, from which we derive valid inequalities {\color{black} which can strengthen the semidefinite relaxation solved at branch-and-bound nodes.}

Our approach improves the scalability of certifiably optimal methods for Problem~\eqref{eq:rco_matrix_completion} compared to the state-of-the-art. In particular, we solve instances of Problem \eqref{eq:rco_matrix_completion} with {\color{black} $\max\{m, n\}$ in the thousands and $k$ up to $5$} to provable optimality in minutes or hours, while previous attempts {\color{black}such as} \cite{MPCO_2021} solve problems where $n=m=50$ and $k=1$ 
in hours, but return optimality gaps larger than $100\%$ after hours when $n\geq 60$ or $k>1$. This is because our approach involves a custom branching scheme that supports imposing semidefinite constraints at the root node and refining relaxations via {\color{black} eigenvector branching}, while \cite{MPCO_2021} used a commercial solver which does not support semidefinite relaxations at the root node and refines relaxations using weaker McCormick disjunctions. 

\subsection{Structure} 
The rest of the paper is structured as follows:

In Section \ref{sec:disj}, we derive Problem \eqref{eq:rco_matrix_completion}'s matrix perspective relaxation and propose {\color{black}a disjunctive branching scheme} for improving it. 
With our disjunctive inequalities, we separate an optimal solution to a relaxation from its feasible region via a single {\color{black}eigenvector disjunction into $2^k$ subregions}. In contrast, using McCormick disjunctions, we prove in Section \ref{ssec:mccormick} that disjuncting {\color{black} over less than $ 2^{n-4}$ subregions} never separates a solution from a McCormick relaxation.

{\color{black}
In Section \ref{sec:bb}, we design a spatial branch-and-bound algorithm that converges to a certifiably optimal solution of 
Problem \eqref{eq:rco_matrix_completion}. 
We also discuss different aspects of our algorithmic implementation, including node selection, branching strategy, and an alternating minimization algorithm to obtain high-quality feasible solutions.
}

In Section \ref{sec:valid.ineq}, we leverage a characterization of the rank of a matrix via its determinant minors to develop a novel convex relaxation for low-rank problems, and derive new valid inequalities. 
{\color{black}
These valid inequalities can be used to tighten the semidefinite relaxations solved at each node in our branch-and-bound scheme.
}

In Section \ref{sec:numres}, we investigate the performance of our branch-and-bound scheme via a suite of numerical experiments. {We identify that our new convex relaxation {\color{black} and valid inequalities} 
reduce (sometimes substantially) the optimality gap at the root node} and that our branch-and-bound scheme solves instances of Problem \eqref{eq:rco_matrix_completion} to certifiable (near) optimality when $\max \{m, n\} = 2500$ in minutes or hours. We also verify that running our branch-and-bound method with a time limit of minutes or hours yields matrices with an out-of-sample MSE up to $50\%$ lower than that of state-of-the-art heuristics, such as the method of \cite{burer2003nonlinear} and its recent refinements. 

\subsection{Notations}
We let non-boldface characters, lowercase bold-faced characters, and uppercase bold-faced characters (e.g., $b, \bm{x}, \bm{A}$) denote scalars, vectors, and matrices, respectively. Calligraphic uppercase characters such as $\mathcal{Z}$ denote sets. We let $[n]$ denote the running set of indices $\{1, \dots, n\}$. We let $\bm{e}$ denote the vector of ones, $\bm{0}$ the vector of all zeros, $\bm{e}_j$ the $j$-th vector of the canonical basis, and $\mathbb{I}$ the identity matrix. We let $\mathcal{S}^n$ denote the set of $n \times n$ symmetric matrices, and $\mathcal{S}^n_+$ denote $n \times n$ positive semidefinite matrices.
We let 
$\projs{n}{k} := \Set{ \bm{Y} \in \mathcal{S}^n: \bm{Y}^2 = \bm{Y}, \ \trace{\bm{Y}} \leq k}$ denote the set of $n \times n$ orthogonal projection matrices with rank at most $k$. Its convex hull is 
$\convhull{\projs{n}{k}} = \Set{\bm{Y} \in S^n : \bm{0} \preceq \bm{Y} \preceq \mathbb{I}, \ \trace{\bm{Y}} \leq k}$.

\section{Mixed-Projection Formulations, Relaxations, and Disjunctions}\label{sec:disj}
In this section, we derive Problem \eqref{eq:rco_matrix_completion}'s semidefinite relaxation in Section \ref{ssec:mprt}, and 
refine its relaxation via {\color{black} eigenvector disjunctions.}
We also establish that our proposed {\color{black} strategy, which we refer to as `eigenvector disjunctions' or `eigenvector branching',} allows us to separate an optimal solution to the original semidefinite relaxation {\color{black}by branching on a single set of eigenvector disjunctions.} In comparison, we justify the poor performance of traditional disjunctions based on McCormick inequalities 
by proving in Section \ref{ssec:mccormick} that {\color{black} McCormick disjunctions with less than $2^{n-4}$} regions cannot improve the semidefinite relaxation. 

\subsection{Mixed-Projection Formulations and Relaxations}\label{ssec:mprt}
Motivated by 
\citet{MPCO_2021,bertsimas2021new}, we introduce a trace-$k$ projection matrix $\bm{Y} \in \mathcal{Y}^k_n$ to model the rank of $\bm{X}$ via the bilinear constraint $\bm{X}=\bm{Y}\bm{X}$. Hence, we can replace the rank constraint on $\bm{X}$ by a linear constraint on $\bm{Y}$: $\mathrm{tr}(\bm{Y}) \leq k$. By \citet[Theorem 1]{bertsimas2021new}, we enforce the bilinear constraint implicitly via the domain of a matrix perspective function and write \eqref{eq:rco_matrix_completion} as:
\begin{align}
    \label{eq:mpco_matrix_completion_mprt}
    \min_{ \bm{Y} \in \projs{n}{k}} \:
    \min_{
        \substack{\bm{X} \in \reals{n \times m} \\ \bm{\Theta} \in \mathcal{S}^m}}
    \quad
    \frac{1}{2\gamma}\trace{\bm{\Theta}}+\frac{1}{2} \sum_{(i,j) \in \mathcal{I}} \left( X_{i,j} - A_{i,j} \right)^2 \ \text{ s.t. } \ 
    \begin{pmatrix}
        \bm{Y} & \bm{X} \\
        \bm{X}^\top & \bm{\Theta}
    \end{pmatrix} \succeq \bm{0}.
\end{align}

By relaxing $\projs{n}{k}$ 
to its convex hull, $\mathrm{Conv}\left(\projs{n}{k}\right) = \{\bm{Y} \in \mathcal{S}^n_+: \bm{Y} \preceq \mathbb{I}, \mathrm{tr}(\bm{Y})\leq k\}$, 
we immediately have the following semidefinite relaxation 
\citep[see also][Lemma 4]{MPCO_2021}:
\begin{align}
    \label{eq:mpco_matrix_completion_mprt_relaxed}
    \min_{
        \substack{\bm{Y} \in \mathrm{Conv}\left(\projs{n}{k}\right) \\ \bm{U} \in \mathbb{R}^{n \times k}}
    } \:
    \min_{
        \substack{\bm{X} \in \reals{n \times m} \\ 
        \bm{\Theta} \in \mathcal{S}^m}
    } \quad & 
    \frac{1}{2\gamma} \mathrm{tr}(\bm{\Theta})+\frac{1}{2} \sum_{(i,j) \in \mathcal{I}} \left( X_{i,j} - A_{i,j} \right)^2
    \quad 
    \text{s.t.} \  
    & \begin{pmatrix}
        \bm{Y} & \bm{X} \\
        \bm{X}^\top & \bm{\Theta}
    \end{pmatrix} \succeq \bm{0},\  \bm{Y} \succeq \bm{U}\bm{U}^\top,
\end{align}
where we introduce the redundant matrix variable $\bm{U} \in \mathbb{R}^{n \times k}$ in order to obtain a tractable set of disjunctions in the next section. 
Note that if $\bm{Y}$ is a rank-$k$ projection matrix then we have that $\bm{Y}=\bm{UU^\top}$ for some $\bm{U} \in \mathbb{R}^{n \times k}$. In our relaxation, we impose the 
semidefinite constraint $\bm{Y} \succeq \bm{UU^\top}$, with the idea to enforce the reverse inequality via the disjunctions we are about to derive.

When we solve \eqref{eq:mpco_matrix_completion_mprt_relaxed}, one of two situations occurs. Either we obtain a solution $\bm{\hat Y}$ with binary eigenvalues and the relaxation is tight, or $\bm{\hat Y}$ has strictly fractional eigenvalues. 
In the latter case, we improve the relaxation by separating $\bm{\hat Y}$ from the feasible region using disjunctive programming techniques, as we propose in the rest of this section.

\subsection{Failure of McCormick Disjunctions}\label{ssec:mccormick}
To further justify the need for a new type of disjunction, we emphasize that most commercial spatial branch-and-bound code{\color{black}s} for non-convex quadratic optimization problems rely on McCormick disjunctions \citep{mccormick1976computability}. However, these disjunctions can be ineffective, as shown in a different context by \cite{khajavirad2023strength}. In this section, we support theoretically the ineffectiveness of McCormick disjunctions: For our matrix completion problem, we show that {\color{black} the value of the relaxation \eqref{eq:mpco_matrix_completion_mprt_relaxed} is left unimproved after $2^{n-4}$ McCormick disjunctions}. Our theoretical result echoes the poor numerical performance of McCormick disjunctions we observe in Section \ref{sec:numres}.

Formally, for each  entry of $\bm{U}$, $U_{i,j}$, we start with box constraints $U_{i,j} \in [-1, 1]$ and recursively partition the region into smaller boxes  $U_{i,j} \in [\underline{U}_{i,j}, \overline{U}_{i,j}]$. 
We also introduce variables $V_{i,j_1,j_2}$ to model the product $U_{i,j_1} U_{i,j_2}$. 
The convex hull of the set $\Set{(v, x, y)| v = xy, (x,y) \in [\underline{x}, \overline{x}] \times [\underline{y}, \overline{y}]}$, denoted $\mathcal{M}(\underline{x}, \overline{x}, \underline{y}, \overline{y})$, is given {\color{black} in \citet{mccormick1976computability} as:}
\begin{align*}
\left\{
    (v, x, y)
    \: \left| \:
    \begin{array}{c}
        {\color{black} 
        (x, y) \in [\underline{x}, \overline{x}] \times [\underline{y}, \overline{y}], 
        }
        \\
        \max \{
            \underline{x}y 
            + \underline{y}x 
            - \underline{x}\underline{y}, 
            \overline{x}y 
            + \overline{y}x 
            - \overline{x}\overline{y}
        \}
        \leq v
        \leq \min \{
            \underline{y}x 
            + \overline{x}y 
            - \overline{x}\underline{y}, 
            \overline{y}x
            + \underline{x}y 
            - \underline{x}\overline{y}
        \} 
    \end{array}
    \right.
\right\}.
\end{align*}
For our matrix completion problem, this yields a disjunction over relaxations of the form: 
\begin{equation} \label{eq:mpco_matrix_completion_mprt_relaxed_U2}
\begin{aligned}
    \min_{
        \substack{
            \bm{Y} \in \mathrm{Conv}(\projs{n}{k}), \\
            \bm{U} \in \reals{n \times k}, \\
            \bm{V} \in \reals{n \times k \times k}
        }
    } 
    & \min_{
        \bm{X} \in \reals{n \times m}, 
        \bm{\Theta} \in \mathcal{S}^m
    } \ 
    \frac{1}{2\gamma}\trace{\bm{\Theta}} 
    +\frac{1}{2} \sum_{(i,j) \in \mathcal{I}} \left( X_{i,j} - A_{i,j} \right)^2
    \quad \text{s.t.} \quad
    \begin{pmatrix} 
        \bm{Y} & \bm{X} \\ 
        \bm{X}^\top & \bm{\Theta} 
    \end{pmatrix} \succeq \bm{0}, 
    \ \bm{Y} \succeq \bm{U}\bm{U}^\top,
    \\
    & \sum_{i=1}^n V_{i,j,j}=1 
    \ \forall \ j \in [k], 
    \  \sum_{i=1}^n V_{i,j_1,j_2} = 0 
    \ \forall \ j_1, j_2 \in [k]: \ j_1 \neq j_2, 
    \\
    & 
    (V_{i,j_1,j_2}, U_{i,j_1}, U_{i,j_2}) \in \mathcal{M}(\underline{U}_{i,j_1}, \overline{U}_{i,j_1}, \underline{U}_{i,j_2}, \overline{U}_{i,j_2}),
    \ \forall \ i \in [n],\ \forall j_1, j_2 \in [k].
\end{aligned}
\end{equation}

{\color{black}
We now prove that a large family of McCormick disjunctions cannot improve on our root node relaxation (proof deferred to Appendix \ref{ssec:proofmccormick}): }
\begin{proposition}
    \label{prop:mccormickisbad_k}
    Consider Problem \eqref{eq:mpco_matrix_completion_mprt_relaxed_U2} and suppose that for every column $j$ in $\bm{U}$ we have a disjunction $\cup_{t}[\underline{U}_{i,j}^t, \overline{U}_{i,j}^t]$ for each $i$ in some index set $\mathcal{I}(j) \subseteq [n]$, 
    but $[\underline{U}_{i', j}, \overline{U}_{i',j}] = [-1,1]$ for all other indices $i' \notin \mathcal{I}(j)$. {\color{black} Further assume that $| \mathcal{I}(j_1) | \leq n-1$ and $| \mathcal{I}(j_1) \cup \mathcal{I}(j_2)| \leq n-4$ for any $j_1,j_2 \in [k], j_1 \neq j_2$.} Then, Problem \eqref{eq:mpco_matrix_completion_mprt_relaxed_U2} possesses the same optimal objective value as Problem \eqref{eq:mpco_matrix_completion_mprt_relaxed}.
\end{proposition}
Proposition \ref{prop:mccormickisbad_k} reveals that, {\color{black} with McCormick disjunctions, we must expand a number of nodes strictly greater than $2^{n-4}$ before improving upon the root node relaxation---indeed, $|\mathcal{I}(j_1) \cup \mathcal{I}(j_2)| \leq |\cup_{j \in [k]} \mathcal{I}(j) |$ and $|\cup_{j \in [k]} \mathcal{I}(j)|$ is lower than the base-2 logarithm of the number of nodes expanded. In addition, the condition of Proposition \ref{prop:mccormickisbad_k} holds if $|\mathcal{I}(j)| \leq (n-4)/2, \forall j \in [k]$ so we can even construct a disjunction with $2^{k (n-4)/2}$ nodes expanded, yet that does not improve the root node relaxation.}
This behavior will contrast with our eigenvector disjunction, which separates the solution to the root node relaxation with a single cut and often improves (numerically) our root node relaxation immediately.
This observation complements a body of work demonstrating that McCormick relaxations are often dominated by other relaxations in both theory and practice \citep[][]{kocuk2018matrix, khajavirad2023strength}. 
Moreover, it challenges the ongoing practice in commercial non-convex solvers, which use McCormick 
disjunctions. 

\subsection{Improving Relaxations via {\color{black} Eigenvector Branching}}\label{ssec:disjgeneric}
We generalize the lifted approach proposed by \citet{saxena2010convex} for mixed-integer QCQO to optimization over rank-$k$ projection matrices, where $k>1$ is allowed. {\color{black} While \citet{saxena2010convex} use their disjunctions to derive disjunctive cuts and design a cutting-plane approach, we construct a branching scheme based on these disjunctions.}

For $\bm{Y}$ to be a rank-$k$ projection matrix, it should satisfy $\hat{\bm{Y}} \preceq \hat{\bm{U}} \hat{\bm{U}^\top}$ in an optimal solution $(\bm{\hat{Y}}, \hat{\bm{U}}, \bm{\hat{X}}, \bm{\hat{\Theta}})$ of the semidefinite relaxation \eqref{eq:mpco_matrix_completion_mprt_relaxed}. Owing to the constraint $\bm{Y}\succeq \bm{UU^\top}$, this would imply that $\hat{\bm{Y}} = \hat{\bm{U}} \hat{\bm{U}}^\top$, and thus $\hat{\bm{Y}}$ is a rank-$k$ matrix. Since $\mathrm{Span}(\hat{\bm{X}}) \subseteq \mathrm{Span}(\hat{\bm{Y}})$ \citep{bertsimas2021new}, we can conclude that $\bm{\hat{X}}$ solves the original rank-constrained problem~\eqref{eq:rco_matrix_completion}. 
Otherwise, we have $\hat{\bm{Y}} \not\preceq \hat{\bm{U}} \hat{\bm{U}}^\top$, i.e., there exists an $\bm{x} \in \reals{n} : \| \bm{x} \|_2 =1$ such that $\| {\color{black} \hat{\bm{U}}^\top} \bm{x} \|_2^2 < \bm{x}^\top \hat{\bm{Y}} \bm{x}$. Therefore, we would like to impose the (non-convex) inequality $\bm{x}^\top \bm{Y} \bm{x}  \leq  \Vert \bm{U}^\top \bm{x} \Vert^2_2$
via a disjunction. 

Observe that, for any feasible $\bm{U}$, $\| {\color{black} {\bm{U}}^\top} \bm{x} \|_2^2 \leq 1$.
First, we decompose the square-norm term
$\| {\color{black} {\bm{U}}^\top} \bm{x} \|_2^2 = \sum_{j \in {\color{black} [k]}} ( \bm{e}_j^\top {\color{black} {\bm{U}}^\top} \bm{x} )^2$. Then we bound each term by a piecewise-linear upper approximation. 
For any $u_0 \in (-1, 1)$, the function $u \in [-1, 1] \mapsto u^2$ has the following piecewise upper approximation with breakpoint $u_0$:
\begin{align}
    \label{eq:disjunction_2}
    u^2 \leq 
    \begin{cases} 
    - u + u u_0 + u_0 
        & \text{if } u \in [-1,u_0], \\
        u + u u_0 - u_0 
        & \text{if } u \in [u_0,1].
    \end{cases}
\end{align}
Using $u_0 = \bm{e}_j{\color{black}^\top} \hat{\bm{U}}^\top \bm{x}$ yields the following upper bound:
\begin{align*}
    ( \bm{e}_j^\top {\color{black} \bm{U}^\top} \bm{x} )^2
    \leq \begin{cases}
        \bm{x}^\top  \hat{\bm{U}} {\color{black} \bm{U}^\top}  \bm{x} 
        + \bm{e}_j^\top(\hat{\bm{U}} - \bm{U}){\color{black}^\top} \bm{x}
        & \quad \text{if } \bm{e}_j^\top {\color{black} \bm{U}^\top} \bm{x} \in [-1, \bm{e}_j^\top {\color{black} \hat{\bm{U}}^\top} \bm{x}],
        \\ 
        \bm{x}^\top \hat{\bm{U}} {\color{black} \bm{U}^\top} \bm{x} + \bm{e}_j^\top(\bm{U} - \hat{\bm{U}}){\color{black}^\top} \bm{x}
        & \quad \text{if } \bm{e}_j^\top {\color{black} \bm{U}^\top} \bm{x} \in [\bm{e}_j^\top {\color{black} \hat{\bm{U}}^\top} \bm{x} , 1].
    \end{cases}
\end{align*}\vspace{-5mm}
and the following disjunction over $2^k$ regions:
\begin{align}
    \label{disjunction2k}
    \bigvee_{L \subseteq {\color{black} [k]}} 
    \Set{
        \bm{Y}
        \: \left| \: 
        \begin{array}{rll}
            \bm{e}_j^\top {\color{black} \bm{U}^\top} \bm{x} &\in [-1, \bm{e}_j^\top {\color{black} \hat{\bm{U}}^\top} \bm{x}]
            & \hspace{-2em} \forall \ j \in L, \\
            \bm{e}_j^\top {\color{black} \bm{U}^\top} \bm{x} &\in [\bm{e}_j^\top {\color{black} \hat{\bm{U}}^\top} \bm{x} , 1] 
            & \hspace{-2em} \forall \ j \in \indexes{k} \setminus L, \\
            \displaystyle
             \bm{x}^\top \bm{Y} \bm{x}
            & \leq 
            \displaystyle
            \bm{x}^\top  \hat{\bm{U}} {\color{black} \bm{U}^\top}  \bm{x} \: + \sum_{j \in L} \bm{e}_j^\top(\hat{\bm{U}} - \bm{U}){\color{black}^\top} \bm{x}
             \: + \sum_{j \in {\color{black} [k]} \setminus L} \bm{e}_j^\top(\bm{U} - \hat{\bm{U}}){\color{black}^\top} \bm{x}
        \end{array}
        \right.
    }.
\end{align}
As a result, we can strengthen our convex relaxation \eqref{eq:mpco_matrix_completion_mprt_relaxed} by {\color{black} following} this disjunction
and optimizing over the $2^k$ resulting convex problems. We formalize
this result in the following proposition:
\begin{proposition}
    \label{corr:disj}
    Let $(\bm{\hat{Y}}, \bm{\hat{U}}, \bm{\hat{X}}, \bm{\hat{\Theta}})$ be an optimal solution to \eqref{eq:mpco_matrix_completion_mprt_relaxed} so that  $\| \hat{\bm{U}^{\color{black}\top}} \bm{x} \|_2^2 < \bm{x}^\top \hat{\bm{Y}} \bm{x}$  for some vector $\bm{x} \in \reals{n}$ such that $\Vert \bm{x}\Vert_2=1$. Then, any solution to \eqref{eq:mpco_matrix_completion_mprt_relaxed} with $\bm{Y} = \bm{UU^\top}$ satisfies \eqref{disjunction2k}, but $(\bm{\hat{Y}}, \bm{\hat{U}}, \bm{\hat{X}}, \bm{\hat{\Theta}})$ does not satisfy \eqref{disjunction2k}.
\end{proposition}
\proof{Proof of Proposition \ref{corr:disj}}
Let $\bm{U}, \bm{Y}$ be matrices so that $\bm{U}\bm{U}^\top=\bm{Y}$. We have $\bm{x}^\top\left(\bm{Y}-\bm{U}\bm{U}^\top\right)\bm{x}=0$ for any vector $\bm{x}$, which implies the disjunction is equivalent to requiring that, for each $j$, either $\bm{e}_j^\top \bm{U}^\top \bm{x} \in [-1, \bm{e}_j^\top\hat{\bm{U}}^\top \bm{x}]$ or $\bm{e}_j^\top\bm{U}^\top \bm{x} \in [ \bm{e}_j^\top\hat{\bm{U}}^\top \bm{x}, 1]$, which is trivially true. 

On the other hand, at the point $\bm{U}=\hat{\bm{U}}$, our approximation of the constraint $\bm{x}^\top \bm{Y} \bm{x}  \leq  \Vert \bm{U}^\top \bm{x} \Vert^2_2$ is tight so our disjunction requires $\bm{x}^\top (\bm{\hat{U}} \bm{\hat{U}}^\top - \hat{\bm{Y}}) \bm{x}\geq 0,$ which contradicts $\bm{x}^\top (\bm{\hat{U}} \bm{\hat{U}}^\top - \hat{\bm{Y}}) \bm{x}< 0.$ \hfill\Halmos
\endproof
Proposition \ref{corr:disj} reveals that 
the disjunction \eqref{disjunction2k} in Problem \eqref{eq:mpco_matrix_completion_mprt_relaxed} separates the optimal solution to \eqref{eq:mpco_matrix_completion_mprt_relaxed} 
whenever it is infeasible for the original problem 
\eqref{eq:rco_matrix_completion}.

\begin{remark} For tractability considerations, we work on a lifted version of the semidefinite relaxation \eqref{eq:mpco_matrix_completion_mprt_relaxed}, which involves additional variables $\bm{U} \in \reals{n \times k}$ such that $\bm{Y} = \bm{UU}^\top$. Instead, we could derive a set of disjunctions on $\bm{Y}$ alone by manipulating the constraint $\bm{Y}^2=\bm{Y}$. Unfortunately, while simpler to derive, this results in a set of $2^n$, rather than $2^k$, disjunctions, which is problematic as $k \ll n$ in practical low-rank matrix completion problems. That being said, working on a lifted formulation may create symmetry issues: if $(\bm{Y}, \bm{U}, \bm{X}, \bm{\Theta})$ is a solution to the relaxation, then so is $(\bm{Y}, \bm{U}\bm{\Pi}, \bm{X}, \bm{\Theta})$ for any $k$-by-$k$ permutation matrix $\bm{\Pi}$. 
Hence, Problem \eqref{eq:mpco_matrix_completion_mprt_relaxed}'s lower bound may not actually improve until we apply a disjunction for each permutation of the columns of $\bm{U}$, which is likely computationally prohibitive in practice. 
We alleviate these issues by exploring branching strategies and symmetry-breaking constraints in Section \ref{sec:bb}. 
\end{remark}

\begin{remark} \label{remark:qpieces}
We obtain our $2^k$ disjunction by constructing a piecewise linear upper approximation of $u \in [-1, 1] \mapsto u^2$ with two pieces, i.e., a single breakpoint $u_0$. We can construct tighter approximations by considering more pieces/breakpoints. 
For example, with the two breakpoints $\{ -|u_0|, +|u_0|\}$, we obtain the following 3-piece linear approximation
\begin{equation}
\begin{aligned}
    \label{eq:disjunction_3}
    u^2 \leq \begin{cases}
        - u - u |u_0| - |u_0| & \quad \text{if } u \in [-1, -|u_0|], \\
        u_0^2 & \quad \text{if } u \in [-|u_0|, |u_0|], \\
        u + u |u_0| - |u_0| & \quad \text{if } u \in [|u_0|, 1], 
    \end{cases}
\end{aligned}
\end{equation}
and with the three breakpoints $\{-|u_0|, 0, |u_0|\}$, 
\begin{equation}
\begin{aligned}
    \label{eq:disjunction_4}
    u^2 \leq \begin{cases}
        - u - u |u_0| - |u_0| & \quad \text{if } u \in [-1, -|u_0|], \\
        - u |u_0| & \quad \text{if } u \in [-|u_0|, 0], \\
        u |u_0| & \quad \text{if } u \in [0, |u_0|], \\        
        u + u |u_0| - |u_0| & \quad \text{if } u \in [|u_0|, 1]. 
    \end{cases}
\end{aligned}
\end{equation}
In general, introducing more pieces results in a stronger but more expensive to compute disjunctive bound. Therefore, there is a trade-off between the quality of the piecewise linear upper approximation and the number of convex optimization problems that need to be solved to compute
the corresponding bound, which we explore numerically in Section \ref{sec:numres}.
\end{remark}

\section{\color{black} A Custom Branch-and-Bound Algorithm}
\label{sec:bb}

In this section, we propose a nonlinear branch-and-bound framework to obtain $\epsilon$-optimal and $\epsilon$-feasible solutions to Problem \eqref{eq:mpco_matrix_completion_mprt} by recursively solving its semidefinite relaxation \eqref{eq:mpco_matrix_completion_mprt_relaxed}, 
partitioning the feasible region via the eigenvector disjunctions derived in Section \ref{sec:disj}.
We first provide a high-level description of our approach and relevant pseudocode, before giving an overview of our branch-and-bound design decisions. {\color{black} Finally, we discuss how alternating minimization can be applied in our branch-and-bound tree to discover high-quality incumbent solutions.}

\subsection{Pseudocode and Convergence Guarantees}
\label{ssec:bb.pseudo}

At the root node, we solve the matrix perspective relaxation \eqref{eq:mpco_matrix_completion_mprt_relaxed}. 
Upon solving the relaxation, we obtain a solution $(\bm{\hat{Y}}, \bm{\hat{U}})$ and compute the smallest eigenvector of $\bm{\hat{U}}\bm{\hat{U}}^\top-\bm{\hat{Y}}$, denoted $\bm{x}$. 
If $\bm{x}^\top \left(\bm{\hat{U}}\bm{\hat{U}}^\top-\bm{\hat{Y}}\right)\bm{x} \geq -\epsilon$, then $\bm{\hat{Y}}$ is a projection matrix up to tolerance $\epsilon$ (since $\bm{\hat{Y}} \succeq \bm{\hat{U}} \bm{\hat{U}}^\top$) and solves \eqref{eq:mpco_matrix_completion_mprt} to optimality and $\epsilon$-feasibility. 
Otherwise, we have $\bm{x}^\top \left(\bm{\hat{U}}\bm{\hat{U}}^\top-\bm{\hat{Y}}\right)\bm{x}<-\epsilon$ 
and we apply a disjunction {\color{black} as in} Equation~\eqref{disjunction2k}. 
{\color{black} Each of the $2^k$ pieces of the disjunction defines} 
a new node or subproblem for {\color{black} our branch-and-bound algorithm}, which is appended to the list of open nodes. 

We proceed recursively by selecting a subproblem from the list of open nodes, solving the corresponding convex relaxation, and imposing a new disjunction which creates more nodes, until we identify an $\epsilon$-feasible solution $(\bm{\hat{Y}}, \bm{\hat{U}})$.

We fathom {\color{black} (or prune)} a node in two situations. 
First, if $\lambda_{\min}(\bm{\hat{U}}\bm{\hat{U}}^\top-\bm{\hat{Y}})\geq-\epsilon$ then {\color{black} the relaxed solution is} an $\epsilon$-feasible solution which provides a global upper bound on \eqref{eq:mpco_matrix_completion_mprt}'s optimal value and we need not impose additional disjunctions at this node. 
As we establish in Theorem \ref{thm:convergence}, this is guaranteed to occur at a sufficiently large tree depth for any $\epsilon>0$. 
Second, if the value of the {\color{black} node} relaxation is within $\epsilon$ of our global upper bound, then no solution with the constraints inherited from this node can improve upon an incumbent solution by more than $\epsilon$. 

We describe our branch-and-bound scheme in Algorithm \ref{alg:mpco_matrix_completion_linear_cuts}.
In Algorithm \ref{alg:mpco_matrix_completion_linear_cuts}, we describe a subproblem as a set of constraints $\mathcal{Q}$ and a depth $t$. 
The constraints in $\mathcal{Q}$ stem from the eigenvector cuts described in Section \ref{sec:disj} and are parameterized by the solution at which they were obtained, $(\hat{\bm{U}}, \hat{\bm{Y}})$, the most negative eigenvector $\bm{x}$, and a vector $\bm{z} \in [q]^k$ encoding which one of the $q^k$ regions we are considering. 
For conciseness, we denote {\color{black}by} $\overline{f}_{i; q}( \cdot \,;\, u_0)$ the $i$-th piece of the piecewise linear upper-approximation with $q$ pieces of $u \mapsto u^2$, obtained from the breakpoint $u_0$.

To verify that Algorithm \ref{alg:mpco_matrix_completion_linear_cuts} {\color{black}converges}, we prove in Theorem \ref{thm:convergence} that a breadth-first node selection strategy which iteratively minimizes \eqref{eq:mpco_matrix_completion_mprt_relaxed} and imposes a disjunctive cut of the form \eqref{disjunction2k}, according to a most negative eigenvector of $\bm{U}{\color{black} \bm{U}^\top}-\bm{Y}$, an optimal solution to \eqref{eq:mpco_matrix_completion_mprt_relaxed} at the current iterate, eventually identifies an $\epsilon$-feasible solution (where $\lambda_{\min}(\bm{U}\bm{U}^\top -\bm{Y})>-\epsilon$) within a finite number of iterations for any $\epsilon>0$. 
The proof of Theorem \ref{thm:convergence} reveals that any node sufficiently deep in our search tree which contains a feasible solution to the semidefinite relaxation (intersected with appropriate constraints from our disjunctions) gives an $\epsilon$-feasible solution to the original problem. 
Therefore, due to the enumerative nature of branch-and-bound, this result verifies that our approach converges for any search and node selection strategy, possibly with a different number of pieces in each disjunction. 

\begin{theorem}\label{thm:convergence}
Let $(\bm{Y}_L, \bm{U}_L)$ denote a solution generated by the $L$-th iterate of the procedure:
\begin{itemize}
    \item For each $\ell \in \mathbb{N}$, set $(\bm{Y}_\ell, \bm{U}_\ell)$ according to the optimal solution of  \eqref{eq:mpco_matrix_completion_mprt_relaxed}, possibly with disjunctive cuts of the form \eqref{disjunction2k}.
    \item If $\lambda_{\min}(\bm{U}_\ell {\color{black} \bm{U}_\ell^\top} - \bm{Y}_\ell)\geq -\epsilon$ then terminate. 
    \item Else, impose the disjunctive cut \eqref{disjunction2k} in \eqref{eq:mpco_matrix_completion_mprt_relaxed} with the most negative eigenvector $\bm{x}_\ell$ of $(\bm{U}_\ell {\color{black} \bm{U}_\ell^\top} - \bm{Y}_\ell)$: $$\bm{x}_\ell \in \argmin_{\bm{x}\in \mathbb{R}^n: \ \Vert \bm{x}\Vert_2=1} \langle \bm{x}\bm{x}^\top, \bm{U}_\ell\bm{U}_\ell^\top-\bm{Y}_\ell\rangle.$$ 
\end{itemize}

For any $\epsilon>0$ there exists an $L \in \mathbb{N}$ so that $\lambda_{\min}(\bm{U}_L {\color{black}\bm{U}_L^\top} - \bm{Y}_L)\geq -\epsilon$ and $\bm{Y}_L$ is an optimal, $\epsilon$-feasible solution to \eqref{eq:mpco_matrix_completion_mprt}. 
Moreover, suppose we set $\epsilon \rightarrow 0$. 
Then, any limit point of $\{\bm{Y}_\ell\}_{\ell=1}^\infty$ {\color{black} defines an optimal solution to \eqref{eq:mpco_matrix_completion_mprt}}.
\end{theorem}
Note that an ``optimal, $\epsilon$-feasible'' solution refers to a solution with objective value at least as good as any solution to \eqref{eq:mpco_matrix_completion_mprt}, which is $\epsilon$-feasible. This is also known as a ``superoptimal'' solution. 

\proof{\color{black} Proof of Theorem \ref{thm:convergence}} {\color{black} In this proof, for the $\ell$-th iterate ($\bm{Y}_\ell,\bm{U}_{\ell})$, we denote $\bm{U}_{\ell}^t$ the $t$-th column of $\bm{U}_{\ell}$.}
Suppose that at the $L$-th iterate this procedure has not converged. Then, since $(\bm{Y}_L, \bm{U}_L)$ satisfies the disjunction \eqref{disjunction2k} for any $\bm{x}_\ell$ such that $\ell<L$, we have that
\begin{equation}
\begin{aligned}
    \sum_{t=1}^k \langle \bm{x}_\ell^\top \bm{U}_\ell^t, \bm{x}_\ell^\top \bm{U}^t_L\rangle+\vert (\bm{U}_\ell^t-\bm{U}_L^t)^\top \bm{x}_\ell\vert \geq \langle \bm{x}_\ell\bm{x}_\ell^\top, \bm{Y}_L\rangle.
\end{aligned}
\end{equation}
But $(\bm{Y}_\ell, \bm{U}_\ell)$ was not $\epsilon$-feasible, and thus we have that
\begin{equation}
\begin{aligned}
    \langle \bm{x}_\ell\bm{x}_\ell^\top, \bm{Y}_\ell\rangle-\sum_{t=1}^k \langle \bm{x}_\ell^\top \bm{U}_\ell^t, \bm{x}_\ell^\top \bm{U}^t_\ell\rangle >\epsilon.
\end{aligned}
\end{equation}
Adding these inequalities then reveals that
\begin{equation}
\begin{aligned}
    \langle \bm{x}_\ell\bm{x}_\ell^\top, \bm{Y}_\ell-\bm{Y}_L\rangle+\sum_{t=1}^k \langle \bm{x}_\ell^\top \bm{U}_\ell^t, \bm{x}_\ell^\top (\bm{U}_L^t-\bm{U}^t_\ell)\rangle+\sum_{t=1}^k \vert \bm{x}_\ell^\top (\bm{U}_\ell^t-\bm{U}^t_L)\vert >\epsilon.
\end{aligned}
\end{equation}
Next, since $\vert \bm{x}_\ell^\top \bm{U}_\ell^t \vert \leq 1$ by construction, using this identity and taking absolute values allows us to conclude that
\begin{equation}
\begin{aligned}
    \vert \langle \bm{x}_\ell\bm{x}_\ell^\top, \bm{Y}_\ell-\bm{Y}_L\rangle\vert+2\sum_{t=1}^k \vert \bm{x}_\ell^\top (\bm{U}_\ell^t-\bm{U}^t_L)\vert >\epsilon.
\end{aligned}
\end{equation}
Moreover, by applying the Cauchy-Schwarz inequality to both terms in this inequality we obtain $\langle \bm{x}_\ell \bm{x}_\ell^\top, \bm{Y}_\ell-\bm{Y}_L\rangle \leq \Vert \bm{Y}_L-\bm{Y}_\ell\Vert_F$ and $\sum_{t=1}^k \vert \bm{x}_\ell^\top (\bm{U}_\ell^t-\bm{U}^t_L)\vert \leq \sum_{t=1}^k \Vert \bm{U}_\ell^t-\bm{U}_L^t\Vert_2=\Vert \bm{U}_\ell-\bm{U}_L\Vert_{2,1}$, {\color{black}where $\Vert \bm{X}\Vert_{2,1}:=\sum_{j \in [k]}\left(\sum_{i \in [n]}X_{i,j}^2\right)^{1/2}$ denotes the $\Vert \cdot \Vert_{2,1}$ norm}. Further, by norm equivalence we have $\Vert \bm{U}_\ell-\bm{U}_L\Vert_{2,1} \leq {\color{black}\sqrt{k}}\Vert \bm{U}_\ell-\bm{U}_L\Vert_F$. Combining these results yields
\begin{equation}
\begin{aligned}
    \Vert \bm{Y}_\ell-\bm{Y}_L\Vert_F+2{\color{black}\sqrt{k}}\Vert \bm{U}_\ell-\bm{U}_L\Vert_F>\epsilon.
\end{aligned}
\end{equation}
{\color{black} That is, with respect to the norm $n_k : (\bm{Y}, \bm{U}) \mapsto \| \bm{Y} \|_F + 2 \sqrt{k} \| \bm{U} \|_F$, our procedure never visits any ball of radius $\frac{\epsilon}{2}$ twice.}
Moreover, the set of feasible $(\bm{Y}, \bm{U})$ is bounded via {\color{black}$\mathbb{I}\succeq \bm{Y}\succeq \bm{U}\bm{U}^\top, \mathrm{tr}(\bm{Y})\leq {k}$}. Therefore, we have that $\Vert \bm{Y}\Vert_F \leq \sqrt{k}$ and $\Vert \bm{U}\Vert_F \leq \sqrt{k}$. {\color{black} Thus, $n_k(\bm{Y}, \bm{U}) \leq \sqrt{k}+2k$} and there are finitely many non-overlapping balls of radius $\frac{\epsilon}{2}$ which contain a point in the feasible region. Therefore, for any $\epsilon>0$, {\color{black} there exists some $L \in \mathbb{N}$ such that} our procedure converges within $L$ iterations. 

{\color{black} 
Finally, let $(\bm{Y}^\star,\bm{U}^\star)$ be a limit point of the sequence $\{(\bm{Y}_\ell, \bm{U}_\ell)\}_{\ell=1}^\infty$, and $(\bm{X}^\star,\bm{\Theta}^\star)$ be the additional variables in the associated optimal solution.  
In the limit, we must have $\bm{Y}^\star = \bm{U}^\star \bm{U}^{\star \top}$ so $\operatorname{rank}(\bm{Y}^\star) \leq \operatorname{rank}(\bm{U}^\star) \leq k$. Let us define $\bm{P}$, the orthogonal projection onto $\mathrm{span}(\bm{Y}^\star)$. The matrix $\bm{P} \in \mathcal{Y}^k_n$ because $\operatorname{rank}(\bm{P}) = \operatorname{rank}(\bm{Y}^\star) \leq k$. Observe that $(\bm{Y}^\star,\bm{U}^\star,\bm{X}^\star,\bm{\Theta}^\star)$ solves a relaxation of \eqref{eq:mpco_matrix_completion_mprt}, so it suffices to show that $(\bm{P},\bm{U}^\star,\bm{X}^\star,\bm{\Theta}^\star)$ is feasible for \eqref{eq:mpco_matrix_completion_mprt} and attains the same objective value. Feasibility follows from the fact that $\mathbb{I}\succeq \bm{Y}^\star$ so $\bm{P}\succeq \bm{Y}^\star$. Objective value is obviously unchanged as it depends on $\bm{X}^\star $ and $ \bm{\Theta}^\star$ only.}
\hfill\Halmos
\endproof

\begin{algorithm}
    \caption{Branch-and-bound for low-rank matrix completion {\color{black}Problem \eqref{eq:rco_matrix_completion}} 
    }
    \label{alg:mpco_matrix_completion_linear_cuts}
    \begin{algorithmic}[1] 
            \State Initialize $\mathcal{Q} = \Set{(\emptyset, 0)}$, 
            $Z_{\text{lower}} = -\infty$, $Z_{\text{upper}} = +\infty$;
            \While{$\mathcal{Q}$ is non-empty and $Z_{\text{upper}} - Z_{\text{lower}} > \epsilon$}
                \label{alg1:1}
                \State Retrieve a problem $(\mathcal{D}, t)$ from the queue: $\mathcal{Q} \leftarrow \mathcal{Q} \setminus (\mathcal{D}, t)$;
                \State Solve the following problem, yielding $(\bm{\hat{\Theta}}_{(t+1)}, \bm{\hat{Y}}_{(t+1)}, \bm{\hat{X}}_{(t+1)}, \bm{\hat{U}}_{(t+1)})$ with objective $Z$:
                \begin{equation} \label{eq:mpco_matrix_completion_mprt_relaxed_alg}
                    \begin{aligned}
                        \min_{
                            \substack{
                                \bm{\Theta} \in \mathcal{S}^m \\
                                \bm{Y} \in \mathcal{S}^n \\
                                \bm{X} \in \reals{n \times m} \\
                                \bm{U} \in \reals{n \times k}
                            }
                        } \quad 
                        & \frac{1}{2} \sum_{(i, j) \in \mathcal{I}} (X_{i,j} - A_{i,j})^2 
                        + \frac{1}{2\gamma} \trace{\bm{\Theta}} \\
                        \text{s.t.} \quad 
                        & \begin{pmatrix}
                            \bm{Y} & \bm{X} \\
                            \bm{X}^\top & \bm{\Theta}
                        \end{pmatrix} \succeq \bm{0}, 
                        \quad \bm{0} \preceq \bm{Y} \preceq \mathbb{I}_n,
                        \quad \trace{\bm{Y}} \leq k, 
                        \quad \begin{pmatrix}
                            \bm{Y} & \bm{U} \\
                            \bm{U}^\top & \mathbb{I}_k 
                        \end{pmatrix} \succeq \bm{0}, \\
                        & \begin{drcases}
                            \bm{U_j}^\top \bm{x} \in [b_{z_j}, b_{{z_j}+1}] 
                            \quad \forall \ j \in \indexes{k} \ \\
                            \langle \bm{Y}, \bm{x} \bm{x}^\top \rangle 
                            \leq \sum_{j=1}^{k} \overline{f}_{z_j; q}(\bm{U_j}^\top \bm{x} ;\ \bm{\hat{U}_j}^\top \bm{x})
                            & 
                        \end{drcases}
                        \forall \ (\bm{\hat{U}}, \bm{\hat{Y}}, \bm{x}, \bm{z}) \in \mathcal{D}
                    \end{aligned}
                    \end{equation}\vspace{-5mm}
                    \If{$\bm{\hat{U}}_{(t+1)} \bm{\hat{U}}^\top_{(t+1)} - \bm{\hat{Y}}_{(t+1)} \succeq \bm{0}$} \Comment{{\color{black}If relaxed solution is feasible for Problem \eqref{eq:mpco_matrix_completion_mprt}}}
                        \If{$Z < Z_{\text{upper}}$}
                            \State $ Z_{\text{upper}} \leftarrow Z $
                             and 
                             $(\bm{\Theta}_{\text{opt}}, \bm{Y}_{\text{opt}}, \bm{X}_{\text{opt}}, \bm{U}_{\text{opt}}) \leftarrow (\bm{\hat{\Theta}}_{(t+1)}, \bm{\hat{Y}}_{(t+1)}, \bm{\hat{X}}_{(t+1)}, \bm{\hat{U}}_{(t+1)})$
                        \EndIf
                    \ElsIf{$Z < Z_{\text{upper}}$} \Comment{{\color{black}If node relaxation is not greater than best solution found}}
                        \State Compute (unit-length) $\bm{x}_{(t+1)}$ s.t. $\bm{x}_{(t+1)}^\top \left( \bm{\hat{U}}_{(t+1)} {\bm{\hat{U}}_{(t+1)}}^\top - \bm{\hat{Y}}_{(t+1)} \right) \bm{x}_{(t+1)} < {\color{black}-\epsilon} $;
                        \For{each $\bm{z} \in [q]^k$ } \Comment{{\color{black}Generate $q^k$ subproblems and add them to the queue}}
                        \State $\mathcal{D}_{\bm{z}} := \mathcal{D} \cup 
                            \{(\bm{\hat{U}}_{(t+1)}, \bm{\hat{Y}}_{(t+1)}, \bm{x}_{(t+1)}, \bm{z}) \}$
                        \State $\mathcal{Q} \leftarrow \mathcal{Q} \cup \Set{ (\mathcal{D}_{\bm{z}}, t+1)} $
                        \EndFor  
                        \State Update $Z_{\text{lower}}$ as the minimum of the lower bounds of all unexplored nodes.
                    \EndIf
                \EndWhile
            \State \textbf{return} $ (\bm{\Theta}_{\text{opt}}, \bm{Y}_{\text{opt}}, \bm{X}_{\text{opt}}, \bm{U}_{\text{opt}}) $
    \end{algorithmic}
\end{algorithm}

In our branch-and-bound code, we also impose symmetry-breaking constraints on $\bm{U}$ throughout the tree, by noting that $\bm{U}$ represents a set of $k$ orthonormal vectors in $\mathbb{R}^n$. We do so via:
\begin{equation}
    U_{ij} \geq 0, 
    \quad \forall \ j \in [k], 
    \ \forall \ i \in \{ n-k+j, n-k+j+1, \dots, n\}
\end{equation}

We now expound on implementation details of Algorithm \ref{alg:mpco_matrix_completion_linear_cuts} pertaining to node selection, branching strategy, and incumbent selection via alternating minimization.

{\color{black}
\subsection{Branch-and-bound Design Decisions}
We briefly review several design decisions in our branch-and-bound scheme in this section, and refer the interested reader to Section \ref{appendix:BnB-decisions} for a more detailed discussion. We evaluate these design decisions empirically in Section \ref{ssec:bnbdesign}.

\paragraph{Node selection (\ref{appendix:node_selection}):} To decide which node to explore next (line 4 in Algorithm \ref{alg:mpco_matrix_completion_linear_cuts}), 
we consider depth-first search (where nodes are selected in a last-in-first-out manner), breadth-first search (where nodes are selected in a first-in-first-out manner), and best-first search (where the node with the lowest remaining lower bound is selected). We find in our numerical experiments that the best-first strategy performs best empirically.

\paragraph{Branching strategy (\ref{appendix:branching_strategy}):}
As discussed in Remark~\ref{remark:qpieces}, our eigenvector branching for a rank-$k$ matrix generates $q^k$ child nodes per parent node when using a piecewise linear approximation of $u \mapsto u^2$ with $q$ pieces. 
Accordingly, another design decision is the number of pieces $q\geq 2$ which should be used. 
}

{\color{black}
\subsection{Incumbent Selection via Alternating Minimization}\label{ssec:AM}
To generate high-quality feasible solutions, we apply an alternating minimization heuristic at each node. Following \cite{burer2003nonlinear,burer_solving_2006}, we decompose $\bm{X} = \bm{U} \bm{V}$, start with an initial solution $\bm{U}_{\text{initial}}$, and iteratively optimize with respect to $\bm{U}$ and $\bm{V}$. We describe the procedure for our matrix completion problem in Algorithm~\ref{alg:mpco_matrix_completion_altmin}.

Compared with the traditional Burer-Monteiro approach, we identify the $\bm{U}$-factor in the decomposition $\bm{X} = \bm{UV}$ with the $\bm{U}$ variable in our semidefinite relaxation. In particular, we want the columns of $\bm{U}$ to be orthogonal so we impose the necessary constraints: $\| \bm{U}_i \|^2_2 \leq 1,\ \forall i \in [k]$ and $\| \bm{U}_i \pm \bm{U}_j \|^2_2 \leq 2,\ \forall i,j \in [k], i \neq j$. At the root node, we impose no additional constraints. At subsequent nodes, we impose the linear constraints on $\bm{U}$ involved in the node relaxation \eqref{eq:mpco_matrix_completion_mprt_relaxed_alg}, i.e., $\bm{U} \in \mathcal{P}_{\bm U} = \{ \bm{U} \: : \: \bm{U}_j^\top \bm{x} \in [b_{z_j}, b_{z_j+1}], \forall j \in [k],\forall (\bm{x},\bm{z}) \in \mathcal{D}\}$.
\begin{algorithm}
    \caption{Alternating minimization for Problem~\eqref{eq:mpco_matrix_completion_mprt_relaxed}
    }
    \label{alg:mpco_matrix_completion_altmin}
    \begin{algorithmic}[1] 
        \Function{AlternatingMinimization}{$\bm{U}_{\text{initial}}$}
        \State Initialize with $\bm{\hat{U}}^0 \leftarrow \bm{U}_{\text{initial}}$, $t \leftarrow 0$
        \While{not converged}
            \State Update $(\bm{\hat{V}}^{t+1}, \bm{\hat{U}}^{t+1})$ as follows, and increment $t$:
            \vspace{-5mm}
            \begin{align}
                \label{eq:altmin_V}
                \bm{\hat{V}}^{t+1} 
                & = 
                \argmin_{\bm{V} \in \reals{k \times m}}
                \quad
                \frac{1}{2} \sum_{(i, j) \in \mathcal{I}} 
                \left( (\bm{\hat{U}}^{t} \bm{V})_{i,j} - A_{i,j} \right)^2
                + \frac{1}{2 \gamma} \Vert \bm{\hat{U}}^t \bm{V} \Vert^2_F
                \\
                \label{eq:altmin_U}
                \bm{\hat{U}}^{t+1} 
                & = \argmin_{
                    \substack{
                        \bm{U} \in \reals{n \times k},\ 
                        \bm{U} \in \mathcal{P}_{\bm{U}}, \\
                        \Vert \bm{U}_i\Vert_2^2 \leq 1
                        \ \forall i \in [k],
                        \ \Vert \bm{U}_i \pm \bm{U}_j\Vert_2^2 \leq 2 
                        \ \forall i{\color{black} < }j \in [k].
                    }
                } \quad
                \frac{1}{2} \sum_{(i, j) \in \mathcal{I}} 
                \left( (\bm{U} \bm{\hat{V}}^{t+1})_{i,j} - A_{i,j} \right)^2
                + \frac{1}{2 \gamma} \Vert \bm{U} \bm{\hat{V}}^{t+1} \Vert^2_F
            \end{align}\vspace{-15mm}\EndWhile
        \State Let $\bm{X}_{\text{altmin}} \leftarrow \bm{\hat{U}}_{\text{end}} \bm{\hat{V}}_{\text{end}}$, and let $\bm{U}_{\text{altmin}} \leftarrow$ compact SVD of $\bm{X}_{\text{altmin}}$
        \State \textbf{return} $\bm{U}_{\text{altmin}}$
        \EndFunction
    \end{algorithmic}
\end{algorithm}
For the initial solution $\bm{U}_{\text{initial}}$, we do the following:
\begin{itemize}
    \item At the root node, we define $\bm{X}_{\text{initial}}$ by $(X_{\text{initial}})_{i,j} = A_{i,j}$ if $(i,j) \in \mathcal{I}$, and 0 otherwise and define $\bm{U}_{\text{initial}} \in \reals{n \times k}$  as the $\bm{U}$-factor of the rank-$k$ truncated SVD of $\bm{X}_{\text{initial}}$.
    \item At child nodes, letting $(\bm{Y}_{R}, \bm{U}_{R}, \bm{X}_{R}, \bm{\Theta}_{R})$ be the solution to the relaxation~\eqref{eq:mpco_matrix_completion_mprt_relaxed_alg}, we define $\bm{U}_{\text{initial}}$ as the $\bm{U}$-factor of a compact SVD of $\bm{Y}_R$.
\end{itemize}

For tractability considerations, we may not run this alternating minimization procedure at every node. Instead, we randomly decide to run Algorithm \ref{alg:mpco_matrix_completion_altmin}, with a probability of selection that decreases with the tree depth. In Section \ref{sec:numres}, we investigate the trade-off between the improved upper bounds from implementing alternating minimization at certain leaf nodes against its time cost.
}

\section{Convex Relaxations and Valid Inequalities}\label{sec:valid.ineq}
In this section, we use a different characterization of low-rank matrices in terms of their determinant minors to derive a {\color{black} stronger} convex relaxation for low-rank matrix completion problems. 
We also 
use this relaxation to derive valid inequalities that strengthen  \eqref{eq:mpco_matrix_completion_mprt_relaxed}. 
{\color{black}
These relaxations can be used as a drop-in replacement for the semidefinite relaxation \eqref{eq:mpco_matrix_completion_mprt_relaxed_alg} solved in Algorithm~\ref{alg:mpco_matrix_completion_linear_cuts}, and we demonstrate their effectiveness numerically in Section~\ref{ssec:numres.presolve}.
}

Our starting point is the observation that the rank of $\bm{X}$ is fully characterized by the following well-known lemma \citep[as proven in, e.g.,][]{johnson1985matrix}:
\begin{lemma}\label{detminor}
Let $\bm{X}$ be a matrix. Then, the rank of $\bm{X}$ is at most $k$ if and only if all $(k+1) \times (k+1)$ minors of $\bm{X}$ have determinant $0$.
\end{lemma}
For our low-rank matrix completion problems, we decompose $\bm{X}$ into a sum of rank-one matrices, apply Lemma \ref{detminor}'s characterization to each rank-one matrix individually, and combine this characterization with our matrix perspective relaxation to obtain stronger relaxations. 

\subsection{Convex Relaxations}\label{ssec:conv.relax}

We construct our new convex relaxation via the following steps.
First, we decompose $\bm{X}=\sum_{t=1}^k \bm{X}^t$, as a sum of rank-one matrices $\bm{X}^t$. 
Second, for each $t \in [k]$,  we introduce a matrix to model the outer product of each vectorized two-by-two minor of ${\color{black} \bm{X}^t}$ with itself, in particular letting $W^t_{i,j}$ model $(X^t_{i,j})^2$ in all moment matrices. 
Third, we use that each two-by-two minor should have zero determinant to eliminate some moment variables {---by introducing the same variable $V^{3,t}_{(i_1,i_2),(j_1,j_2)}$ on each entry of the anti-diagonal of the moment matrix.} 
Fourth, we model the term $X_{i,j}^2=\left(\sum_{t=1}^k X_{i,j}^t\right)^2=\sum_{t \in [k]}(X_{i,j}^t)^2+2\sum_{t' < t}X_{i,j}^t X_{i,j}^{t'}$ in the objective by also performing a relaxation on the moment matrix generated by the vector $\begin{pmatrix} 1 & X_{i,j}^1 & X_{i,j}^2 & \ldots & X_{i,j}^k\end{pmatrix}$ and selecting the appropriate off-diagonal terms to model each product $X_{i,j}^{t_1} X_{i,j}^{t_2}$. Finally, we replace $X_{i,j}^2$ with $W_{i,j}$ in the objective where applicable and
link $\Theta_{i,j}$ with the appropriate terms in $\bm{W}$ to impose appropriate objective pressure in our relaxation. We have the following result:
\begin{proposition}
\label{prop:lasserrematrixcompletionk}
The following {\color{black}semidefinite} relaxation is at least as strong as the matrix perspective relaxation for matrix completion \eqref{eq:mpco_matrix_completion_mprt_relaxed} where $k \in [\min(n, m)]$:
\begin{align}
    \label{prob:rankk_relax_slice}
     \min_{
        \substack{
            \bm{X}^t, \bm{W}, \bm{W}^t \in \reals{n \times m} \ \forall t \in [k], \\
            \bm{Y} \in \mathrm{Conv}(\mathcal{Y}^{k}_n), \
            \bm{\Theta} \in \mathcal{S}^m_+, \\ \bm{V}^{\cdot,t}, \bm{H}^{t,t'}, \forall t, t' \in [k]
        }
    } &
    \frac{1}{2\gamma}\mathrm{tr}(\bm{\Theta})+\frac{1}{2}\sum_{(i,j) \in \mathcal{I}}\left(\color{black}A_{i,j}^2+W_{i,j}-2 \sum_{t \in [k]} X_{i,j}^t A_{i,j}\right)\\
     \text{\rm s.t.} \quad  & {\begin{pmatrix} 
            \bm{Y} 
            & \sum_{t \in [k]}\bm{X}^t \\ 
            \sum_{t \in [k]}{\bm{X}^t}^\top 
            & \bm{\Theta} 
        \end{pmatrix} \succeq \bm{0}}, \nonumber\\
        & 
        {\small
        \begin{pmatrix}
            1           & X^t_{i_1,j_1}                 & X^t_{i_1,j_2}                 & X^t_{i_2,j_1}                 & X^t_{i_2,j_2}                 \\[1ex]
            X^t_{i_1,j_1} & W^t_{i_1,j_1}                 & V_{i_1,(j_1,j_2)}^{1,t}       & V_{(i_1,i_2),j_1}^{2,t}       & V_{(i_1,i_2),(j_1,j_2)}^{3,t} \\[1ex]
            X^t_{i_1,j_2} & V_{i_1,(j_1,j_2)}^{1,t}       & W^t_{i_1,j_2}                 & V_{(i_1,i_2),(j_1,j_2)}^{3,t} & V_{(i_1,i_2),j_2}^{2,t}       \\[1ex]
            X^t_{i_2,j_1} & V_{(i_1,i_2),j_1}^{2,t}       & V_{(i_1,i_2),(j_1,j_2)}^{3,t} & W^t_{i_2,j_1}                 & V_{i_2,(j_1,j_2)}^{1,t}       \\[1ex]
            X^t_{i_2,j_2} & V_{(i_1,i_2),(j_1,j_2)}^{3,t} & V_{(i_1,i_2),j_2}^{2,t}       & V_{i_2,(j_1,j_2)}^{1,t}       & W^t_{i_2,j_2}
        \end{pmatrix} \succeq \bm{0},
        \quad
        \begin{aligned}
            & \forall \ i_1 < i_2 \in [n], \\
            & \forall \ j_1 < j_2 \in [m], \\
            & \forall \ t \in [k],
        \end{aligned}
        }\nonumber
        \\
        & \begin{pmatrix}
            1           & X_{i,j}^{1}  & \ldots & X_{i,j}^{k}   \\[1ex]
            X_{i,j}^{1} & W_{i,j}^{1}   & \ldots & H_{i,j}^{1,k} \\[1ex]
            \vdots      & \vdots              & \ddots & \vdots        \\[1ex]
            X_{i,j}^{k} & H_{i,j}^{1,k}  & \ldots & W_{i,j}^{k}
        \end{pmatrix} \succeq \bm{0},
        \quad \forall \ i \in [n], \ j \in [m],\nonumber
        \\
        & \Theta_{{j,j}} = \sum_{i \in [n]} W_{i,j}, \ W_{i,j} = \sum_{t \in [k]} \left( W_{i,j}^t + \sum_{t' \in [k]: t'\neq t} H_{i,j}^{t',t} \right) 
        \ \forall \ i \in [n], \ j \in [m].\nonumber
\end{align}
\end{proposition}

\proof{Proof of Proposition \ref{prop:lasserrematrixcompletionk}}
    This follows immediately from the fact that any feasible solution to \eqref{prob:rankk_relax_slice} is feasible in \eqref{eq:mpco_matrix_completion_mprt_relaxed} with objective value {\color{black}no larger, since $W_{i,j}\geq X_{i,j}^2$}. \hfill\Halmos
\endproof



\subsection{Partial Convex Relaxations}\label{ssec:valid.ineqs}
Despite its strength, 
our new relaxation is also computationally prohibitive to solve at scale. 
Accordingly, we use it to derive valid inequalities that we further impose {\color{black}on} \eqref{eq:mpco_matrix_completion_mprt_relaxed} to develop more scalable (yet less tight) semidefinite relaxations. 

To develop our relaxation, we begin with the following problem, which is equivalent to \eqref{eq:mpco_matrix_completion_mprt_relaxed} but also includes each slice $\bm{X}^t$ and variables $W_{i,j}$ to model $X_{i,j}^2$ for each $(i,j)$:
\begin{equation}\label{prob:rankk_relax_slice2}
\begin{aligned}
\min_{
        \substack{
            \bm{X}^t, \bm{W}, \bm{W}^t \in \reals{n \times m} \ \forall t \in [k], \\
            \bm{Y} \in \mathrm{Conv}(\mathcal{Y}^{k}_n), \
            \bm{\Theta} \in \mathcal{S}^m_+, \bm{V}
        }
    } & \quad 
    \frac{1}{2\gamma}\mathrm{tr}(\bm{\Theta})+\frac{1}{2}\sum_{(i,j) \in \mathcal{I}} \left(\color{black}A_{i,j}^2+W_{i,j}-2 \sum_{t \in [k]}X_{i,j}^tA_{i,j}\right) \\  
    \text{s.t.} & \quad 
    \begin{pmatrix} \bm{Y} & \sum_{t \in [k]}\bm{X}^t \\ \sum_{t \in [k]}{\bm{X}^t}^\top & \bm{\Theta} \end{pmatrix} \succeq \bm{0}, \\
    & \Theta_{{j,j}} = \sum_{i \in [n]} W_{i,j}, \ \forall j \in [m], \ \left(\sum_{t \in [k]}X_{i,j}^t\right)^2 \leq W_{i,j},
    \ \forall \ i \in [n], \ j \in [m]. 
\end{aligned}
\end{equation}
Given a solution to this relaxation, we select a subset of the $2 \times 2$ minors of $\bm{X}$ according to a pre-specified criterion, e.g., all $2 \times 2$ minors where all four entries $(i_1, i_2, j_1, j_2)$ are observed, and a random sample of $2 \times 2$ minors where at least three entries are observed. 
Next, for each such minor $(i_1, i_2, j_1, j_2)$, we introduce appropriately indexed variables $\bm{V}, \bm{W}^t, \bm{H}$, omit the constraint linking $X_{i,j}^t$ and $W_{i,j}$, and impose the following constraints:
\begin{align*}
     & \begin{pmatrix}
        1             & X^t_{i_1,j_1}                 & X^t_{i_1,j_2}                 & X^t_{i_2,j_1}                 & X^t_{i_2,j_2}                 \\[1ex]
        X^t_{i_1,j_1} & W^t_{i_1,j_1}                 & V_{i_1,(j_1,j_2)}^{1,t}       & V_{(i_1,i_2),j_1}^{2,t}       & V_{(i_1,i_2),(j_1,j_2)}^{3,t} \\[1ex]
        X^t_{i_1,j_2} & V_{i_1,(j_1,j_2)}^{1,t}       & W^t_{i_1,j_2}                 & V_{(i_1,i_2),(j_1,j_2)}^{3,t} & V_{(i_1,i_2),j_2}^{2,t}       \\[1ex]
        X^t_{i_2,j_1} & V_{(i_1,i_2),j_1}^{2,t}       & V_{(i_1,i_2),(j_1,j_2)}^{3,t} & W^t_{i_2,j_1}                 & V_{i_2,(j_1,j_2)}^{1,t}       \\[1ex]
        X^t_{i_2,j_2} & V_{(i_1,i_2),(j_1,j_2)}^{3,t} & V_{(i_1,i_2),j_2}^{2,t}       & V_{i_2,(j_1,j_2)}^{1,t}       & W^t_{i_2,j_2}
    \end{pmatrix} \succeq \bm{0}, 
    \quad \forall \ t \in [k],\\
    & 
    \begin{pmatrix}
        1           & X_{i,j}^{1}   & X_{i,j}^{2}   & \ldots & X_{i,j}^{k}   \\[1ex]
        X_{i,j}^{1} & W_{i,j}^{1}   & H_{i,j}^{1,2} & \ldots & H_{i,j}^{1,k} \\[1ex]
        X_{i,j}^{2} & H_{i,j}^{1,2} & W_{i,j}^{2}   & \ldots & H_{i,j}^{2,k} \\[1ex]
        \vdots      & \vdots        & \vdots        & \ddots & \vdots        \\[1ex]
        X_{i,j}^{k} & H_{i,j}^{1,k} & H_{i,j}^{2,k} & \ldots & W_{i,j}^{k}   \\
    \end{pmatrix} \succeq \bm{0},
    \quad \forall \ i \in \{i_1, i_2\}, \ j \in \{j_1, j_2\}, \\
    & W_{i,j} = \sum_{t \in [k]} \left(
        W_{i,j}^t
        + \sum_{t' \in [k]: t' \neq t} H_{i,j}^{t',t}
    \right), 
    \quad \forall \ i \in \{i_1, i_2\}, \ j \in \{j_1, j_2\}.
\end{align*}
All in all, we impose $k$ $5 \times 5$ PSD constraints and at most 4 $(k+1) \times (k+1)$ PSD constraints for each $2 \times 2$ submatrix we aim to model, which allows us to control the computational complexity of our relaxation via the number of cuts at each problem size. 
In particular, the computational complexity of modeling a Shor relaxation of a given $2 \times 2$ minor scales independently of $n, m$.


\section{Numerical Experiments}\label{sec:numres}
In this section, we evaluate the numerical performance of our branch-and-bound scheme,
implemented in \verb|Julia| version {\color{black} 1.9} using \verb|Mosek| version 10.0 to solve all semidefinite optimization problems. 
All experiments were conducted on MIT's Supercloud cluster \citep{reuther2018interactive}, which hosts Intel Xeon Platinum 8260 processors. 
To perform our experiments, we generate synthetic instances of matrix completion problems, as described in Section \ref{ssec:synthetic}. 

We evaluate the effectiveness of the valid inequalities from Section \ref{sec:valid.ineq} 
in Section \ref{ssec:numres.presolve}. 
Next, in Section \ref{ssec:bnbdesign}, we benchmark the performance of different node selection, branching, breakpoint, and incumbent generation strategies for our 
algorithm. 
Finally, we evaluate the scalability of our branch-and-bound algorithm
in Section \ref{ssec:scalability}, in terms of its ability to identify an optimal solution, and find feasible solutions that outperform a {\color{black} popular} alternating minimization strategy. 
To bridge the gap between theory and practice, we make our branch-and-bound implementation available at \url{github.com/sean-lo/OptimalMatrixCompletion.jl} and our experiments at \url{github.com/sean-lo/OLRMC_experiments}.

\subsection{Generation of Synthetic Instances} \label{ssec:synthetic}
We compute a matrix of observations, $\bm{A}_{\text{full}} \in \reals{n \times m}$, from a low-rank model: 
$\bm{A}_{\text{full}} = \bm{U} \bm{V} + \epsilon \, \bm{Z}$,
where the entries of $\bm{U} \in \reals{n \times k}, \bm{V} \in \reals{k \times m}$, and $\bm{Z} \in \reals{n \times m}$ are drawn independently from a standard normal distribution, and $\epsilon \geq 0$ models the degree of noise. We fix $\epsilon = 0.1$. 
We then sample a random subset $\mathcal{I} \subseteq [n] \times [m]$ of predefined size, which contains at least one entry in each row and column of the matrix \citep[see also][Section 1.1.2]{candes2009exact}. 
To do so, if the target size $|\mathcal{I}|$ is large enough, we iteratively add random entries until this property is satisfied, which will happen with {\color{black}fewer} than $|\mathcal{I}|$ draws with high probability. 
In regimes where $|\mathcal{I}|$ is small, i.e., close to $k (n + m)$, we use a random permutation matrix to sample $k (n + m)$ entries, one in each row and column, directly. 
We then independently sample the remaining entries of $\mathcal{I}$ to reach the desired size.

For simplicity, in our experiments, we set $m=n$ and only vary the dimension of the problem, $n$, and the number of observed entries, $|\mathcal{I}|$. 
Also, to allow for a better comparison across instances of varying size $n$, we generate one large $N \times N$ matrix $\bm{A}_{\text{full}}$ and consider its top-left $n$-by-$n$ submatrices for various values of $n$, which creates correlations between the instances generated for different values of $n$—the instances generated for a given size $n$, however, remain independent since they come from different matrices $\bm{A}_{\text{full}}$ obtained with different random seeds. 

We remark that this data generation process is standard in the literature and generates matrices from the same distribution as those considered by many other authors \citep[e.g.,][]{candes2009exact, MPCO_2021}.

\subsection{Root Node: Strengthened Relaxations}\label{ssec:numres.presolve}
In this section, we evaluate the benefit of the valid inequalities presented in Section \ref{sec:valid.ineq}. 

As argued in Section \ref{ssec:valid.ineqs}, we can strengthen the matrix perspective relaxation \eqref{eq:mpco_matrix_completion_mprt_relaxed} by imposing additional semidefinite constraints (Shor LMIs) on all $2 \times 2$ minors of the slices of $\bm{X}$, $\bm{X}^t, t\in [k]$. 
To minimize the impact on tractability, however, we do not impose these constraints for all minors. 
Instead, we consider two-by-two minors with all four entries present in $\mathcal{I}$ (denoted $\mathcal{M}_4$), or with at least three entries in $\mathcal{I}$ ($\mathcal{M}_3$). 
We compare the original relaxation \eqref{eq:mpco_matrix_completion_mprt_relaxed} with \eqref{prob:rankk_relax_slice2} strengthened with Shor LMIs for all minors in $\mathcal{M}_4$, all minors in $\mathcal{M}_4$ and a (random) half of the minors in $\mathcal{M}_3$, and all minors in $\mathcal{M}_4$ and $\mathcal{M}_3$, using two metrics: optimality gap at the root node (the upper bound is obtained via alternating minimization) and computational time for solving the relaxation.

\begin{figure}[h!]
\centering
\begin{subfigure}{.5\textwidth}
    \centering
    \includegraphics[width=\linewidth]{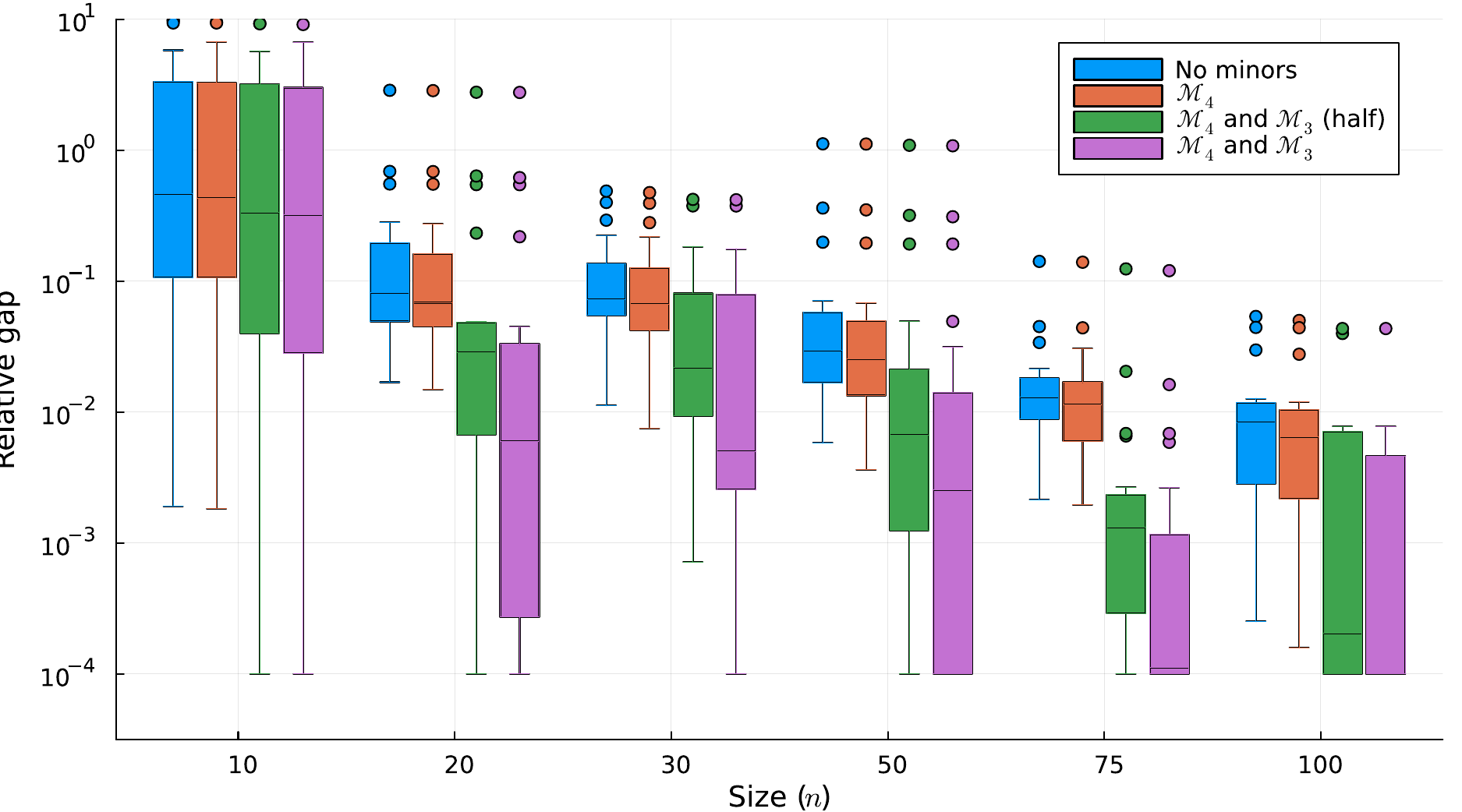}
    \caption{Root node relative gap}
    \label{fig:mc1_root_boxplot_loglb_smalllarge}
\end{subfigure}%
\begin{subfigure}{.5\textwidth}
    \centering
    \includegraphics[width=\linewidth]{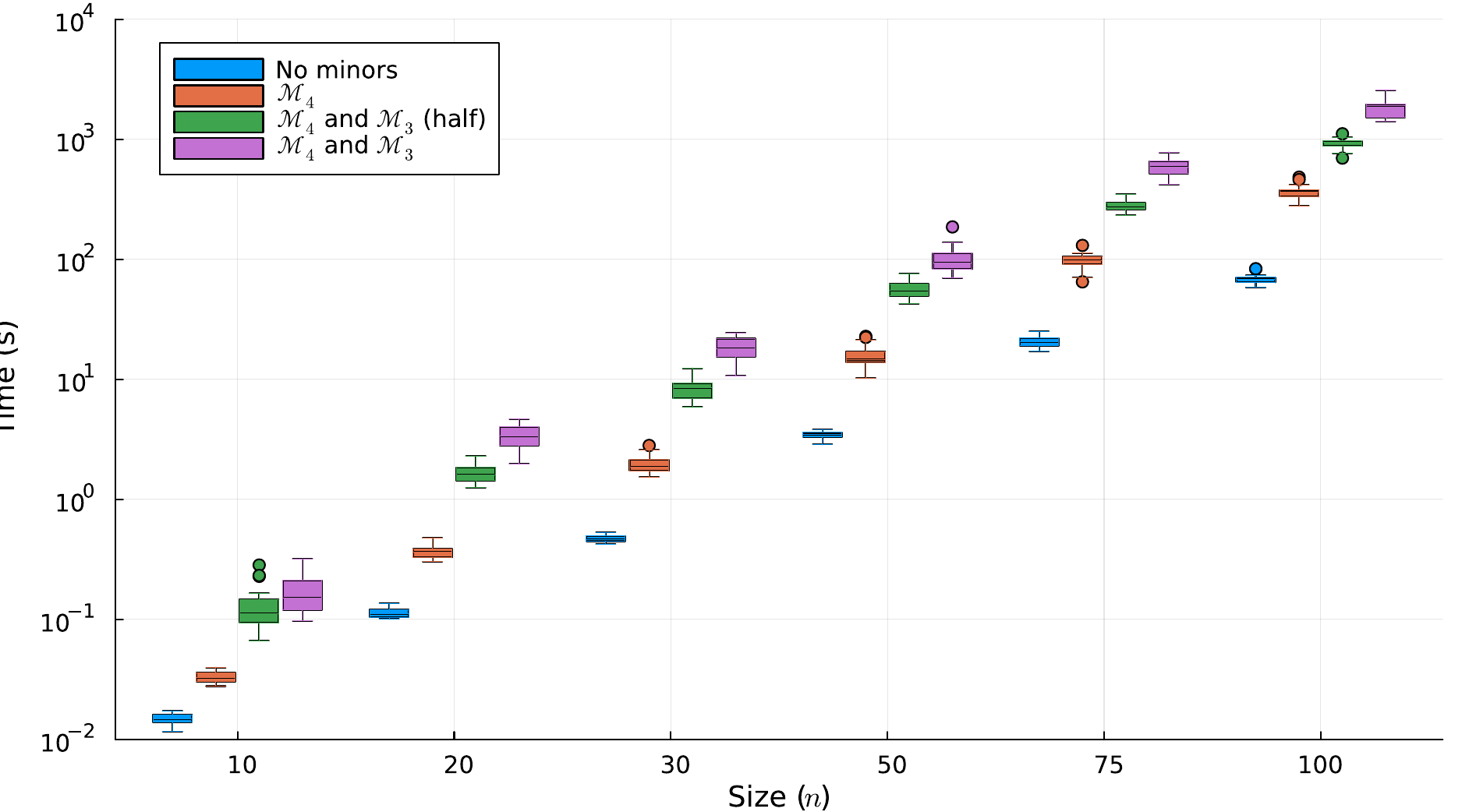}
    \caption{Computation time (s)}
    \label{fig:mc1_root_boxplot_time_smalllarge}
\end{subfigure}
\caption{{\color{black} Root node relative gap (a) and computation time (b) at the root node for rank-one $n$-by-$n$ matrix completion problems with $2n \log_{10}(n)$ filled entries, in a regime with low regularization ($\gamma = 80.0$).}}
\label{fig:mc_root_loglb_time}
\end{figure}

We consider rank-one matrix completion where $n \in \{10, 20, 30, 50, 75, 100\}$, and $|\mathcal{I}| = 2n \log(n)$. 
Figures \ref{fig:mc1_root_boxplot_loglb_smalllarge} and \ref{fig:mc1_root_boxplot_time_smalllarge} represent the optimality gap at the root node and computational time respectively of using these strengthened relaxations. 
{\color{black}
We observe that, 
while adding Shor LMIs for minors in $\mathcal{M}_4$ only has a small impact on the optimality gap, considering all minors in $\mathcal{M}_4$ and $\mathcal{M}_3$ significantly reduce the optimality gaps. 
For example, for $n \geq 50$, it reduces the median optimality gap by nearly two orders of magnitude (from $10^{-2}$ down to $10^{-4}$). This comes with a corresponding one-order-of-magnitude increase in computation time. 
Introducing Shor LMIs for all minors in $\mathcal{M}_4$ and half of the minors in $\mathcal{M}_3$ partially alleviates the computational burden while retaining some benefits in relaxation tightness, providing us with greater flexibility to control the optimality-tractability tradeoff. Detailed results are reported in Tables~\ref{tab:mc1_root_solve_time}--\ref{tab:mc1_root_root_node_gap} in \ref{appendix:numres.presolve}. 
}

\begin{figure}[h!]
\centering
\begin{subfigure}{.5\textwidth}
    \centering
    \includegraphics[width=\linewidth]{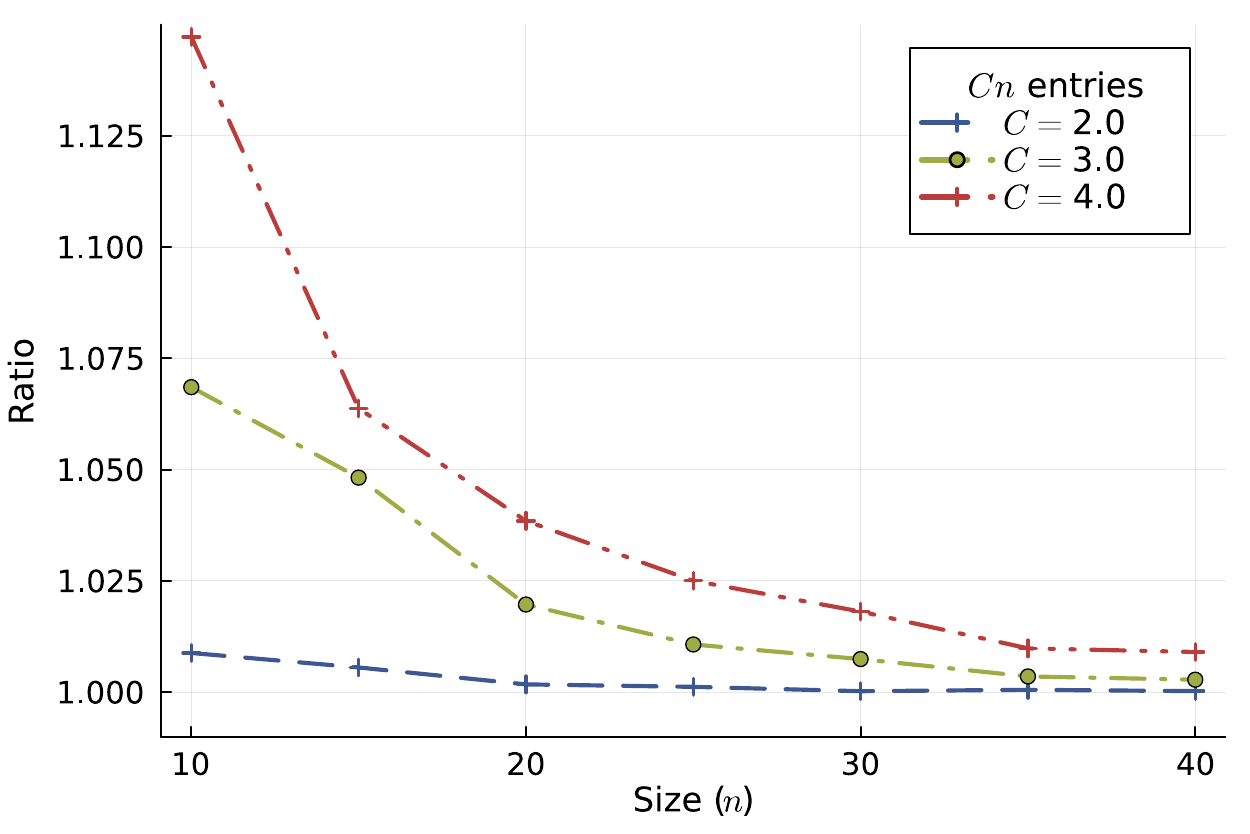}
    \caption{(Scaled) lower bounds of $\mathcal{M}_4$ (rank-one)}
    \label{fig:mc_root_1_ratio_4_none_size}
\end{subfigure}%
\begin{subfigure}{.5\textwidth}
    \centering
    \includegraphics[width=\linewidth]{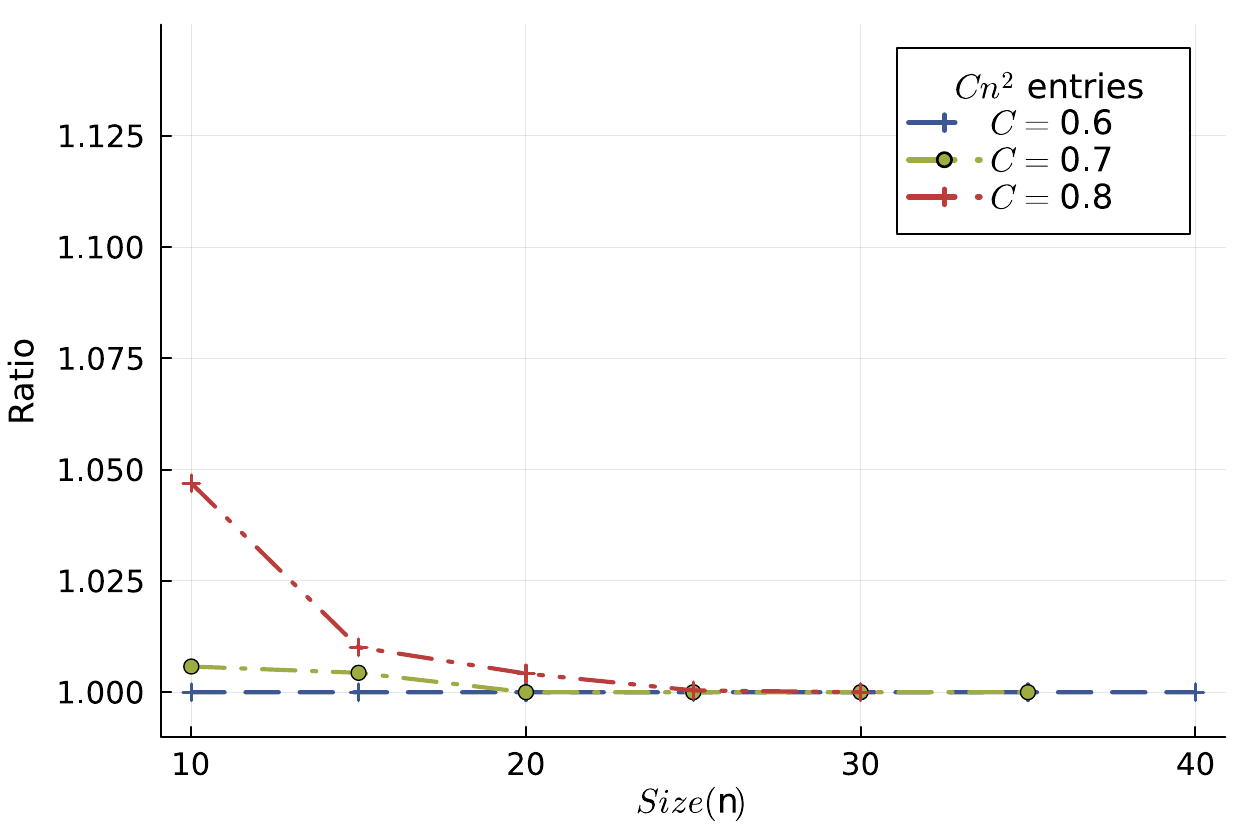}
    \caption{(Scaled) lower bounds of $\mathcal{M}_4$ (rank-two)}
    \label{fig:mc_root_2_ratio_4_none_size}
\end{subfigure}
    \caption{(Scaled) lower bounds of $\mathcal{M}_4$ relaxation relative to the relaxation with no Shor LMIs, with $C n$ or $C n^2$ entries, with low regularization ($\gamma = 80.0$).}
    \label{fig:mc_root_12_ratio_4_none_size}
\end{figure}

{We observe that the relative strength of our relaxation crucially depends on the proportion of missing entries, especially as the rank $k$ increases. 
Figures \ref{fig:mc_root_1_ratio_4_none_size}--\ref{fig:mc_root_2_ratio_4_none_size} 
show the objective value of our relaxation with Shor LMIs for minors in $\mathcal{M}_4$, relative to the relaxation without any minors, for different values of $|\mathcal{I}|$. 
We consider $|\mathcal{I}| = C n$ (resp. $|\mathcal{I}| =C n^2$) for rank-one (resp. rank-two) instances in Figure  \ref{fig:mc_root_1_ratio_4_none_size} (resp. Figure~\ref{fig:mc_root_2_ratio_4_none_size}). 
In both cases, we observe that the relative benefit from adding Shor LMIs to the root node relaxation increases as $C$ increases.
We also observe that the number of entries required to observe an improvement from the Shor LMIs is higher in the rank-two than in the rank-one case---notably in its scaling with respect to $n$. 
A larger number of observed entries, however, might be computationally more expensive (because there are more Shor LMIs to add) and not available in practice. 
}

\subsection{Branch-and-bound Design Decisions}\label{ssec:bnbdesign}
In this section, we benchmark the efficacy of different algorithmic design options for our branch-and-bound algorithm, including whether to use {\color{black}an eigenvector branching scheme} (as developed in Section \ref{sec:disj}) or a {\color{black}McCormick-based branching scheme} (as described in Section \ref{ssec:mccormick}); the node expansion strategy (breadth-first, best-first, or depth-first); and whether alternating minimization, as described in Section \ref{ssec:AM}, should be run at child nodes in the tree. 

Tables \ref{tab:mc1_gap}--\ref{tab:mc1_time} report the final optimality gaps and total computational time of our branch-and-bound scheme with different configurations as we vary $n$. 
We impose two termination criteria for these experiments: a relative optimality gap {\color{black}of} $10^{-4}$ and a time limit of one hour. 
Eigenvector disjunctions achieve optimality gaps about an order of magnitude smaller than McCormick disjunctions. 
Running alternating minimization at child nodes also improves the average optimality gap by around an order of magnitude, which provides evidence that the Burer-Monteiro alternating minimization method is not optimal in practice when run from the root node, and can be improved via branch-and-bound. 
Moreover, the best-first node selection strategy, which comprises selecting the unexpanded node with the smallest lower bound at each iteration, outperforms both breadth-first and depth-first search. This phenomenon is consistent across $p$ and $\gamma$; see Tables \ref{tab:mc1_gap_less_lessreg}--\ref{tab:mc1_time_more_lessreg}. 
Accordingly, we use best-first search with eigenvector disjunctions and alternating minimization run at child nodes throughout the rest of the paper, unless explicitly indicated otherwise.

We remark that in preliminary numerical experiments, we also considered solving matrix completion problems via the multi-tree branch-and-cut approach proposed in \cite{MPCO_2021}, and directly with \verb|Gurobi|'s non-convex QCQP solver. 
Unfortunately, neither approach was competitive with either the McCormick disjunction or the eigenvector disjunction approach, likely because \verb|Gurobi| does not allow semidefinite constraints to be imposed, and the root node relaxation without semidefinite constraints is often quite weak. 
Indeed, for instances where $n=50$ and $k>1$, neither approach produced a better lower bound than the matrix perspective relaxation.

{
\begin{table}\footnotesize
    \centering
    \begin{tabular}{S[table-format=2.0] c S[table-format=1.2e-1] S[table-format=1.2e-1] S[table-format=1.2e-1] S[table-format=1.2e-1] S[table-format=1.2e-1] S[table-format=1.2e-1]}
        \toprule 
        & & \multicolumn{3}{c}{With McCormick disjunctions} & \multicolumn{3}{c}{With eigenvector disjunctions} 
        \\
        \cmidrule(r){3-5} \cmidrule(l){6-8}
        {\multirow{2}{*}{$n$}} & {Alternating} & {\multirow{2}{*}{Best-first}} & {\multirow{2}{*}{Breadth-first}} & {\multirow{2}{*}{Depth-first}} & {\multirow{2}{*}{Best-first}} & {\multirow{2}{*}{Breadth-first}} & {\multirow{2}{*}{Depth-first}} 
        \\
        & {minimization} & & & & & & 
        \\
        \midrule
        10 & \xmark & 6.63e-01 & 6.82e-01 & 6.84e-01 & 5.72e-01 & 5.61e-01 & 2.65e-01 \\ 
        10 & \cmark & 1.26e-02 & 2.18e-02 & 1.63e-01 & \cellcolor[gray]{0.9}3.57e-03 & 1.01e-02 & 5.29e-02 \\ 
        \midrule
        20 & \xmark & 8.24e-02 & 8.24e-02 & 8.24e-02 & 6.44e-02 & 6.75e-02 & 8.78e-02 \\ 
        20 & \cmark & 9.44e-03 & 1.19e-02 & 1.19e-02 & \cellcolor[gray]{0.9}1.71e-03 & 3.36e-03 & 9.52e-03 \\ 
        \midrule
        30 & \xmark & 4.88e-02 & 4.88e-02 & 4.88e-02 & 4.15e-02 & 4.23e-02 & 4.79e-02 \\ 
        30 & \cmark & 1.36e-02 & 1.71e-02 & 1.71e-02 & \cellcolor[gray]{0.9}1.71e-03 & 2.88e-03 & 1.12e-02 \\ 
        \midrule
        40 & \xmark & 4.92e-02 & 4.92e-02 & 4.92e-02 & 4.44e-02 & 4.49e-02 & 2.14e-02 \\ 
        40 & \cmark & 2.03e-02 & 2.03e-02 & 2.03e-02 & \cellcolor[gray]{0.9}6.53e-04 & 9.50e-04 & 1.10e-02 \\ 
        \midrule
        50 & \xmark & 1.50e-02 & 1.50e-02 & 1.50e-02 & 2.68e-02 & 2.81e-02 & 2.97e-02 \\ 
        50 & \cmark & 6.06e-03 & 6.06e-03 & 6.09e-03 & \cellcolor[gray]{0.9}2.22e-04 & 3.36e-04 & 8.83e-03 \\ 
        \bottomrule
    \end{tabular}
    \vspace{1pt}
    \caption{Final optimality gap 
    across rank-one matrix completion problems with $|\mathcal{I}| = pn \log_{10}(n)$ filled entries, averaged over 20 instances per row ($p = 2.0$, $\gamma = 20.0$).}
    \label{tab:mc1_gap}
\end{table}
}

{
\begin{table}\footnotesize
    \centering
    \begin{tabular}{
        S[table-format=2.0] c 
        S[table-format=4.2] S[table-format=4.2] S[table-format=4.2] 
        S[table-format=4.2] S[table-format=4.2] S[table-format=4.2]
    }
        \toprule 
        & & \multicolumn{3}{c}{With McCormick disjunctions} & \multicolumn{3}{c}{With eigenvector disjunctions} 
        \\
        \cmidrule(r){3-5} \cmidrule(l){6-8}
        {\multirow{2}{*}{$n$}} & {Alternating} & {\multirow{2}{*}{Best-first}} & {\multirow{2}{*}{Breadth-first}} & {\multirow{2}{*}{Depth-first}} & {\multirow{2}{*}{Best-first}} & {\multirow{2}{*}{Breadth-first}} & {\multirow{2}{*}{Depth-first}} 
        \\
        & {minimization} & & & & & & 
        \\
        \midrule
        10 & \xmark & 3061.5 & 3078.2 & 3240.0 & 2448.9 & 2546.7 & 3240.0 \\ 
        10 & \cmark & \cellcolor[gray]{0.9}1453.9 & 1603.7 & 3061.8 & 1675.8 & 1974.4 & 2824.6 \\ 
        \midrule
        20 & \xmark & 3060.1 & 3060.1 & 3060.1 & 1939.4 & 2254.9 & 2894.2 \\ 
        20 & \cmark & 2540.1 & 2700.1 & 2700.1 & \cellcolor[gray]{0.9}1539.7 & 1862.5 & 2370.4 \\ 
        \midrule
        30 & \xmark & 3600.3 & 3600.4 & 3600.2 & 3084.1 & 3155.5 & 3600.3 \\ 
        30 & \cmark & 3060.4 & 3060.4 & 3060.3 & \cellcolor[gray]{0.9}2285.9 & 2363.7 & 3060.4 \\ 
        \midrule
        40 & \xmark & 2881.3 & 2881.1 & 2881.0 & 1818.1 & 1827.6 & 2881.4 \\ 
        40 & \cmark & 2701.1 & 2701.2 & 2701.2 & \cellcolor[gray]{0.9}1203.4 & 1290.2 & 2701.0 \\ 
        \midrule
        50 & \xmark & 2703.0 & 2702.9 & 2702.8 & 1749.5 & 2400.8 & 2703.7 \\ 
        50 & \cmark & 2343.4 & 2343.1 & 2343.1 & \cellcolor[gray]{0.9}959.32 & 1299.7 & 2437.2 \\ 
        \bottomrule
    \end{tabular}
    \vspace{1pt}
    \caption{Total computational time (s)
    across rank-one matrix completion problems with $| \mathcal{I} | = pn \log_{10}(n)$ filled entries, averaged over 20 instances per row ($p = 2.0$, $\gamma = 20.0$).}
    \label{tab:mc1_time}
\end{table}
}

\begin{figure}[h]
\centering
\begin{subfigure}{.5\textwidth}
    \centering
    \includegraphics[width=0.8\linewidth]{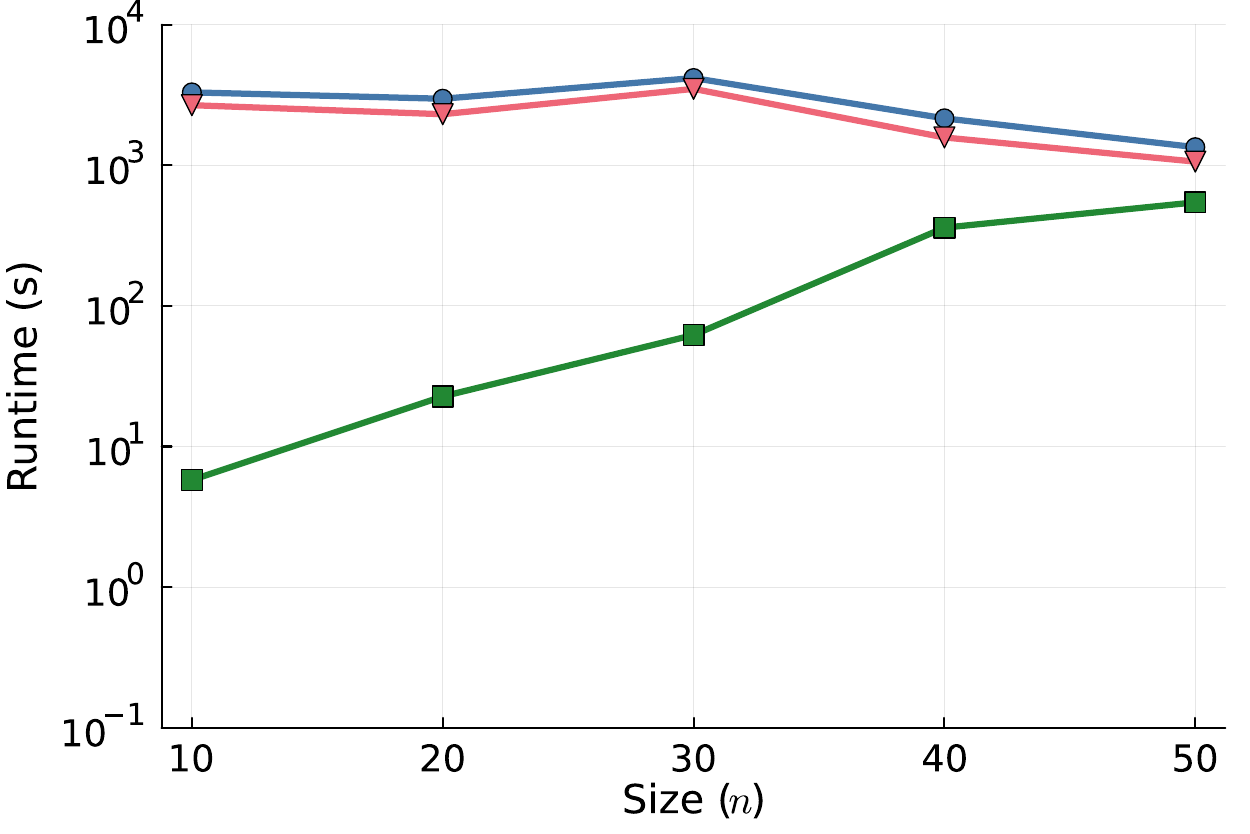}
    \label{fig:mc1_branch_less_morereg}
\end{subfigure}%
\begin{subfigure}{.5\textwidth}
    \centering
    \includegraphics[width=0.8\linewidth]{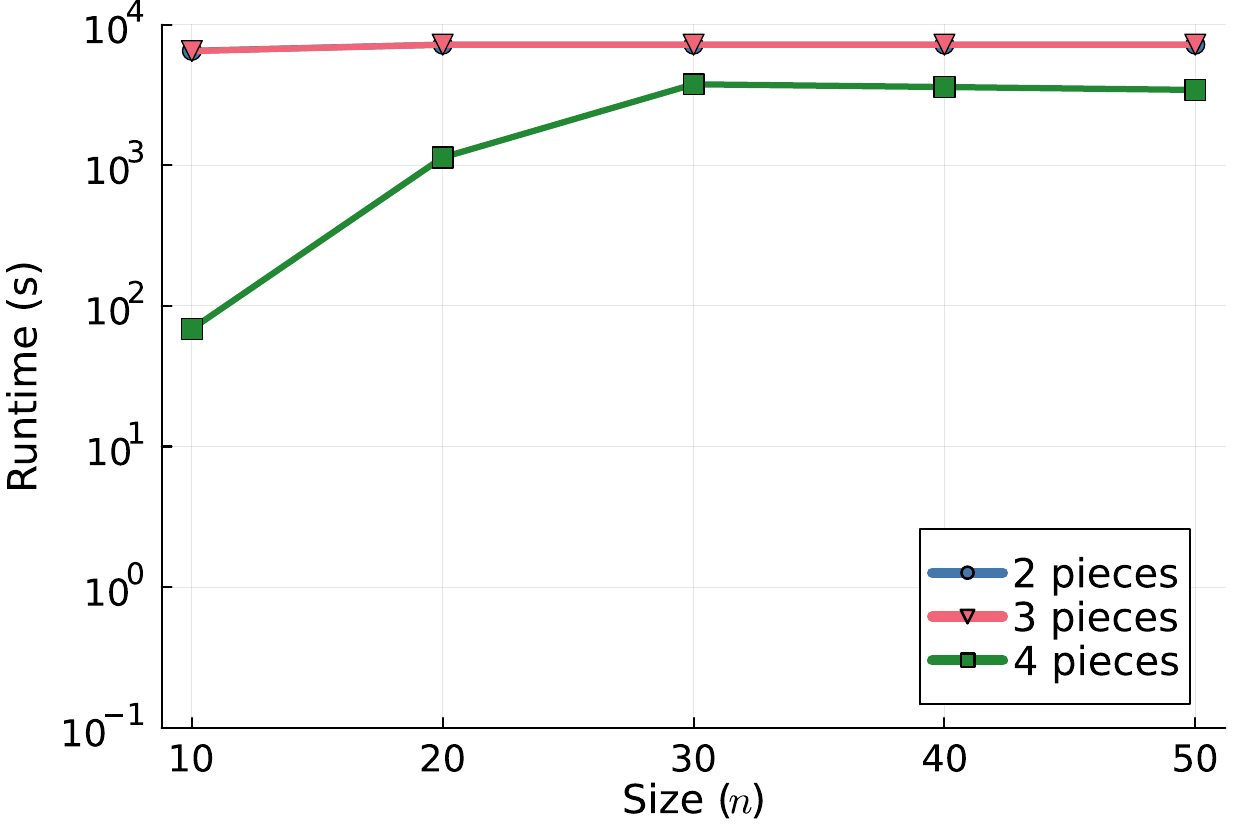}
    \label{fig:mc1_branch_less_lessreg}
\end{subfigure}
\begin{subfigure}{.5\textwidth}
    \centering
    \includegraphics[width=0.8\linewidth]{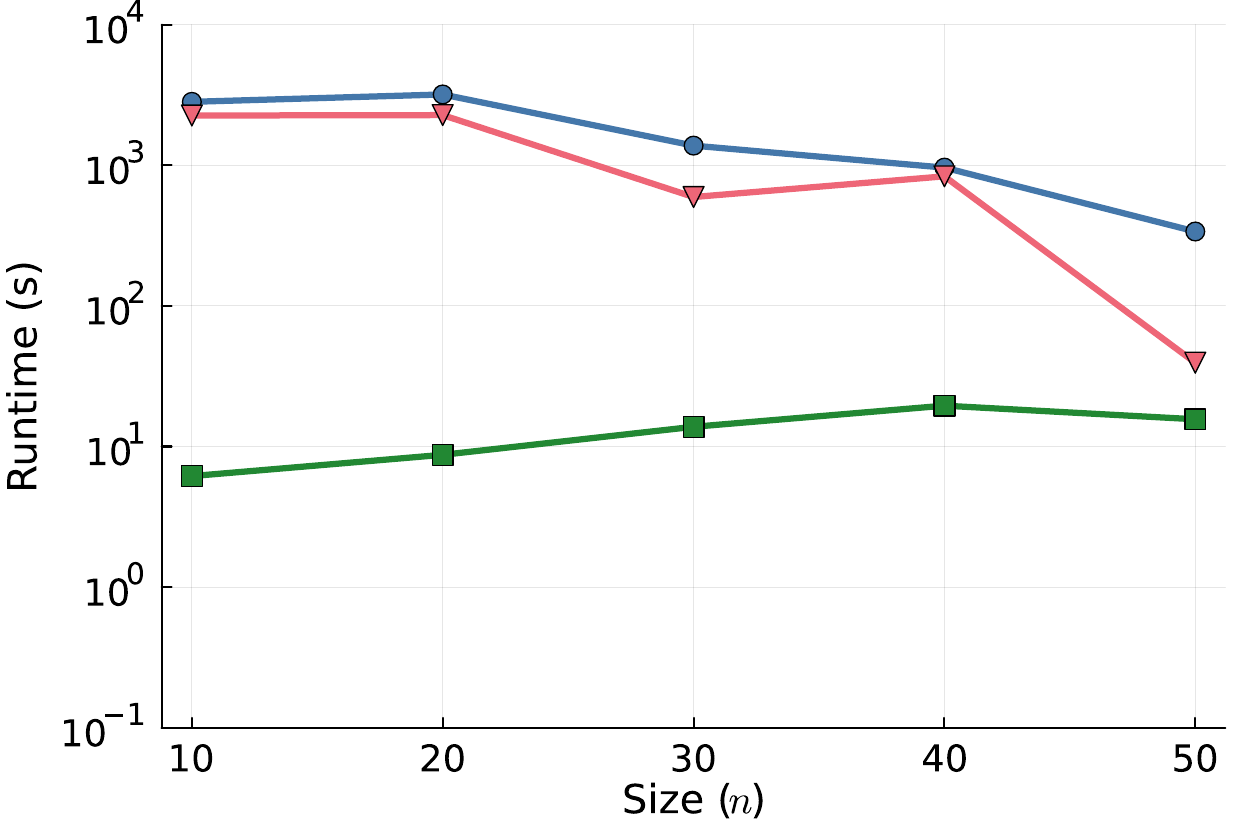}
    \label{fig:mc1_branch_more_morereg}
\end{subfigure}%
\begin{subfigure}{.5\textwidth}
    \centering
    \includegraphics[width=0.8\linewidth]{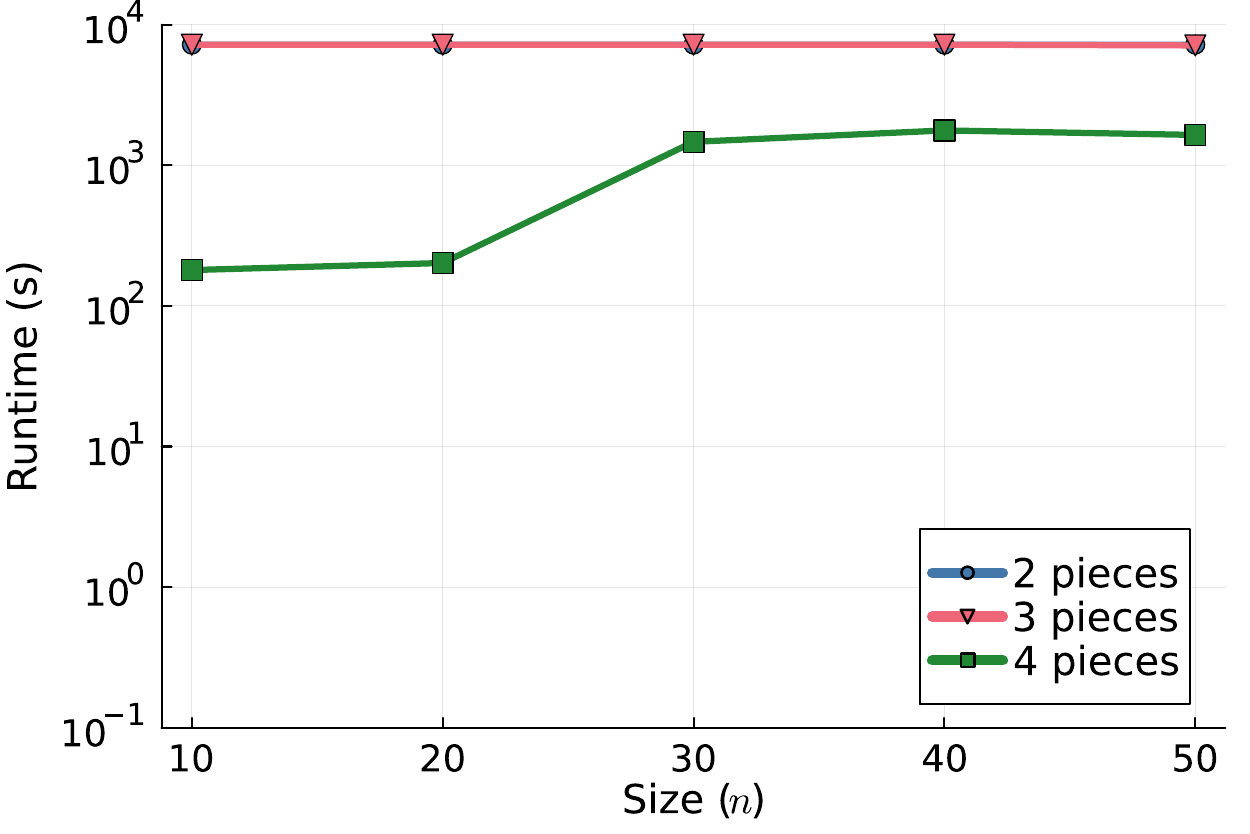}
    \label{fig:mc1_branch_more_lessreg}
\end{subfigure}
\caption{Comparison of time taken to optimality (relative gap $10^{-4}$) for rank-one matrix completion problems with $pn \log_{10}(n)$ filled entries, over different numbers of pieces $q \in \{2, 3, 4\}$ in {\color{black}the} upper-approximation. We set $p=2.0$ (less entries) in the top plots, and $p=3.0$ (more entries) in the bottom plots. We set $\gamma=20.0$ (more regularization) in the left plots, and $\gamma=80.0$ (less regularization) in the right plots.}
\label{fig:mc1_branch}
\end{figure}

We also investigate the effect of the number of pieces in our disjunction, $q$, 
on the time needed to achieve a $10^{-4}$ optimality gap for rank-one matrix completion problems, in Figure~\ref{fig:mc1_branch}. 
We do not observe a significant difference between using $q=2$ or $q=3$ pieces. 
However, we find that implementing a disjunctive scheme with $q=4$ pieces allows our branch-and-bound strategy to converge orders of magnitude faster, across all values of $p$ and $\gamma$. 
We suspect this occurs because four-piece disjunctions include zero as a breakpoint, which breaks some symmetry. 
However, we also observe {\color{black}that} this advantage vanishes as $n$ increases; Figure~\ref{fig:mc1_size_gap_pknlog_linear_linear3} confirms this on larger instances.

\subsection{Scalability Experiments}\label{ssec:scalability}

Table~\ref{tab:mc1_time} reveals that the strongest implementation of our algorithm solves rank-one matrix completion problems where $m = n = 50$ in minutes. Accordingly, we now investigate the scalability of our algorithm with this configuration and its ability to obtain higher-quality low-rank matrices than heuristics. We apply our algorithm to rank-$k$ matrix completion problems on rectangular instances of size $50 \times m$, with $k m \log_{10}(m)$ observed entries, $k \in \{1, 2, 3, 4, 5\}$, and a time limit of three hours. 

\begin{figure}
\centering
\begin{subfigure}{.5\textwidth}
    \centering
    \includegraphics[width=\linewidth]{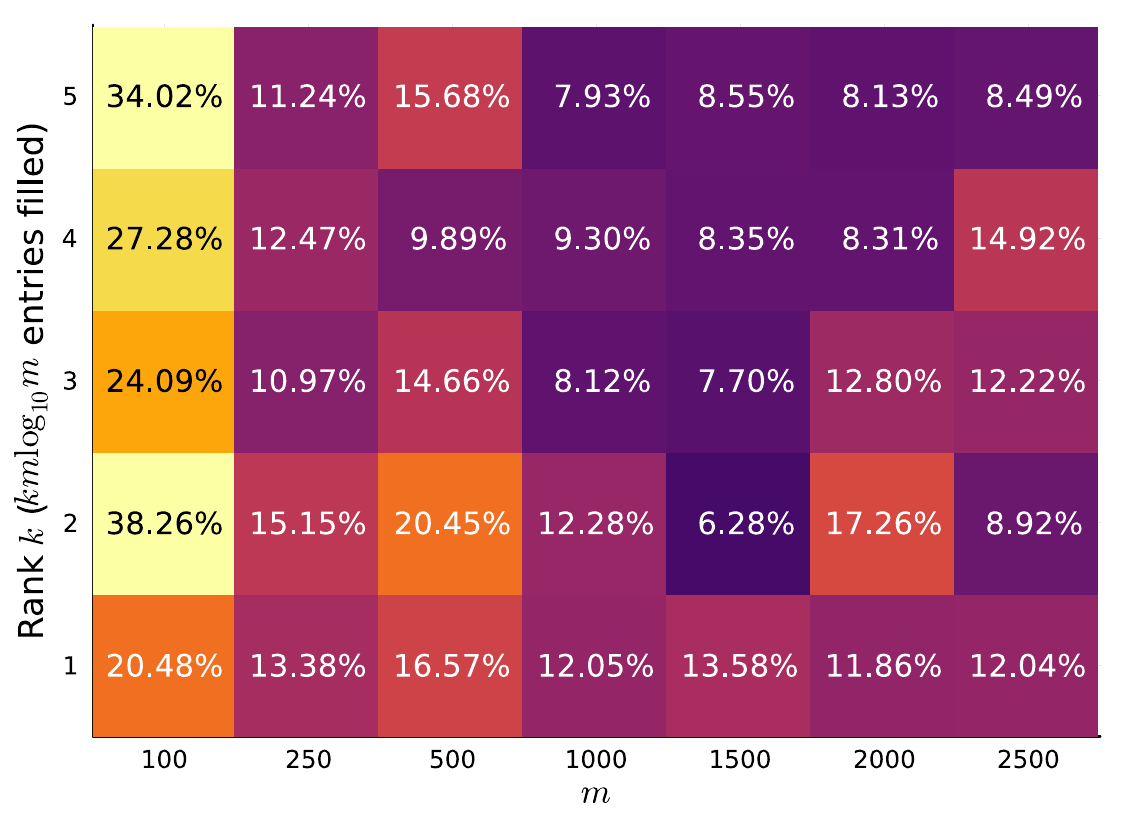}
    \label{fig:mc_scale_relative_gap_root_node_mean_heatmap}
\end{subfigure}%
\begin{subfigure}{.5\textwidth}
    \centering
    \includegraphics[width=\linewidth]{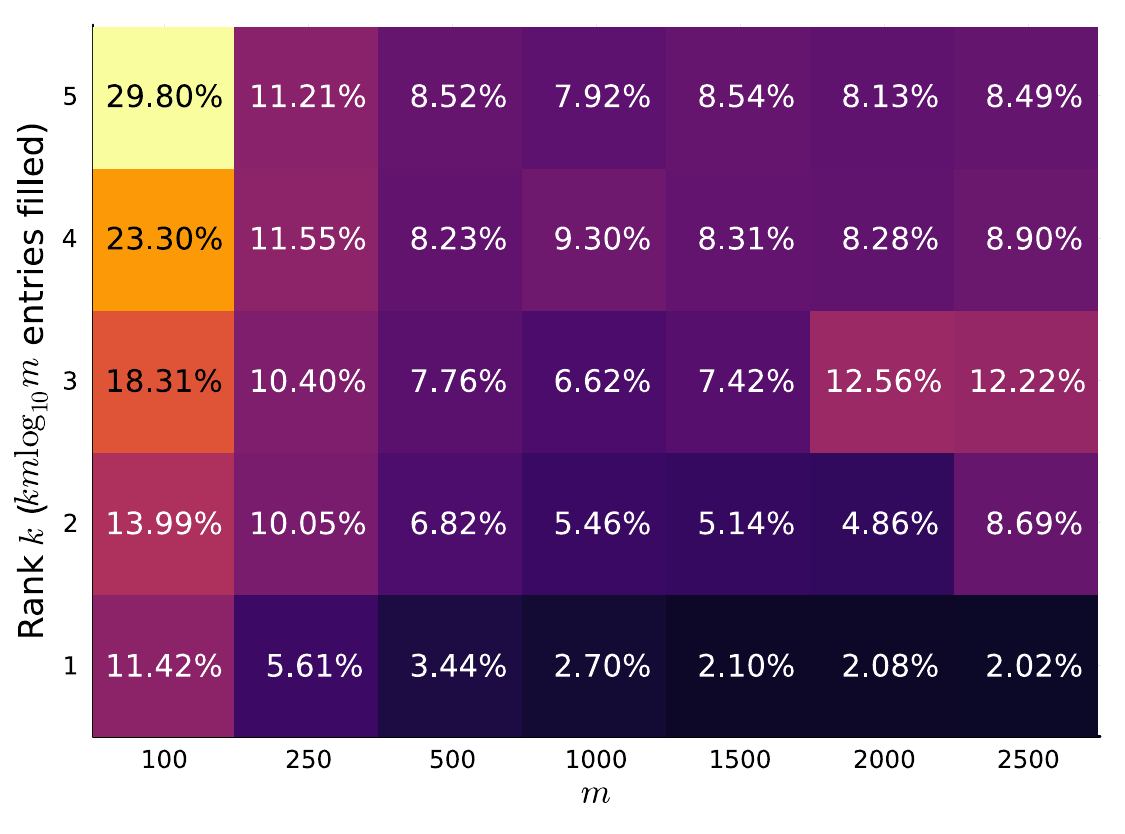}
    \label{fig:mc_scale_relative_gap_mean_heatmap}
\end{subfigure}
\caption{Comparison of relative optimality gap at root node (left) and after running branch-and-bound for three hours (right) for rank-$k$ matrix completion problems of dimension $50 \times m$, with $k m \log_{10}(m)$ filled entries, varying $m$ and $k$, with $\gamma = 120.0$, averaged over 10 random instances.}
\label{fig:mc_scale_relative_gap_heatmap}
\end{figure}

Figure~\ref{fig:mc_scale_relative_gap_heatmap} depicts the relative gap between the root node relaxation and the best incumbent solution at the root node (left panel), and after applying branch-and-bound for three hours (right panel). 
{\color{black}
Comparing the left and right panels, we observe that Algorithm \ref{alg:mpco_matrix_completion_linear_cuts} is effective at reducing the optimality gap across all $(m,k)$ values, and is especially effective for $k = 1$. 
}

\begin{figure}
\centering
\begin{subfigure}{.5\textwidth}
    \centering
    \includegraphics[width=\linewidth]{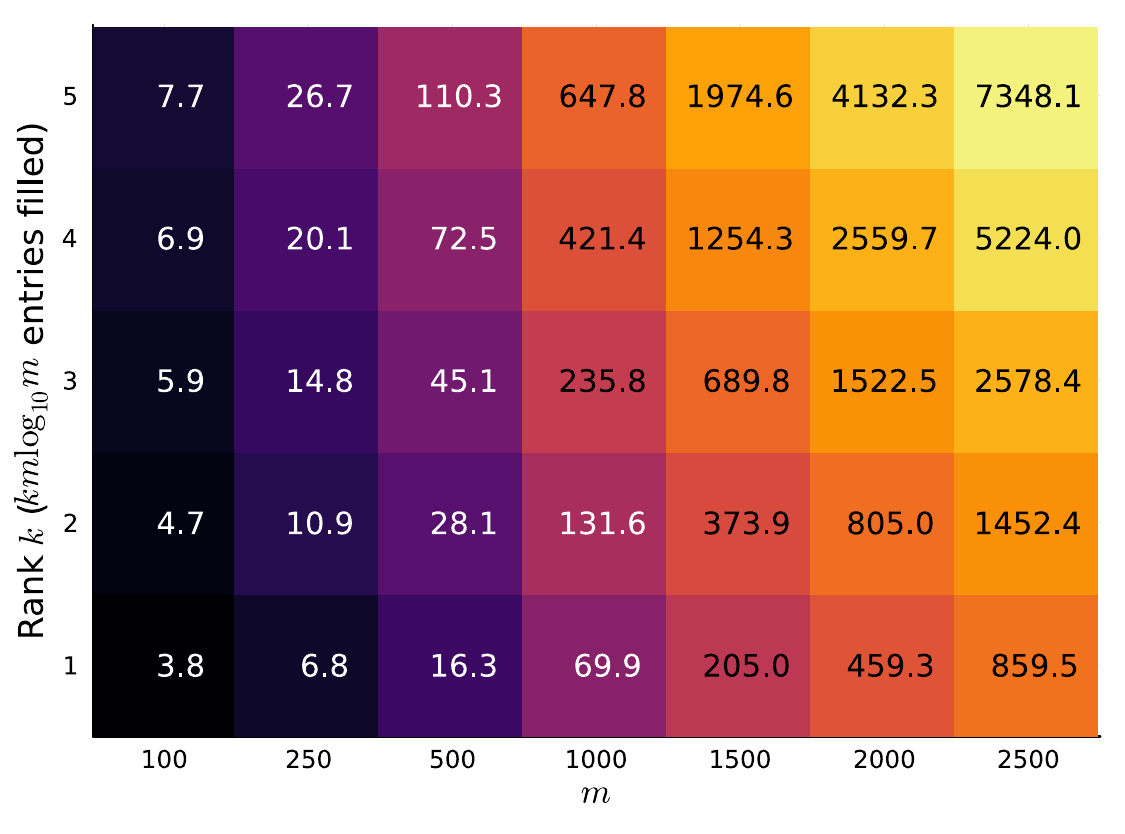}
    \caption{SDP relaxation solve time (in seconds)}
    \label{fig:mc_scale_relaxation_time_heatmap}
\end{subfigure}%
\begin{subfigure}{.5\textwidth}
    \centering
    \includegraphics[width=\linewidth]{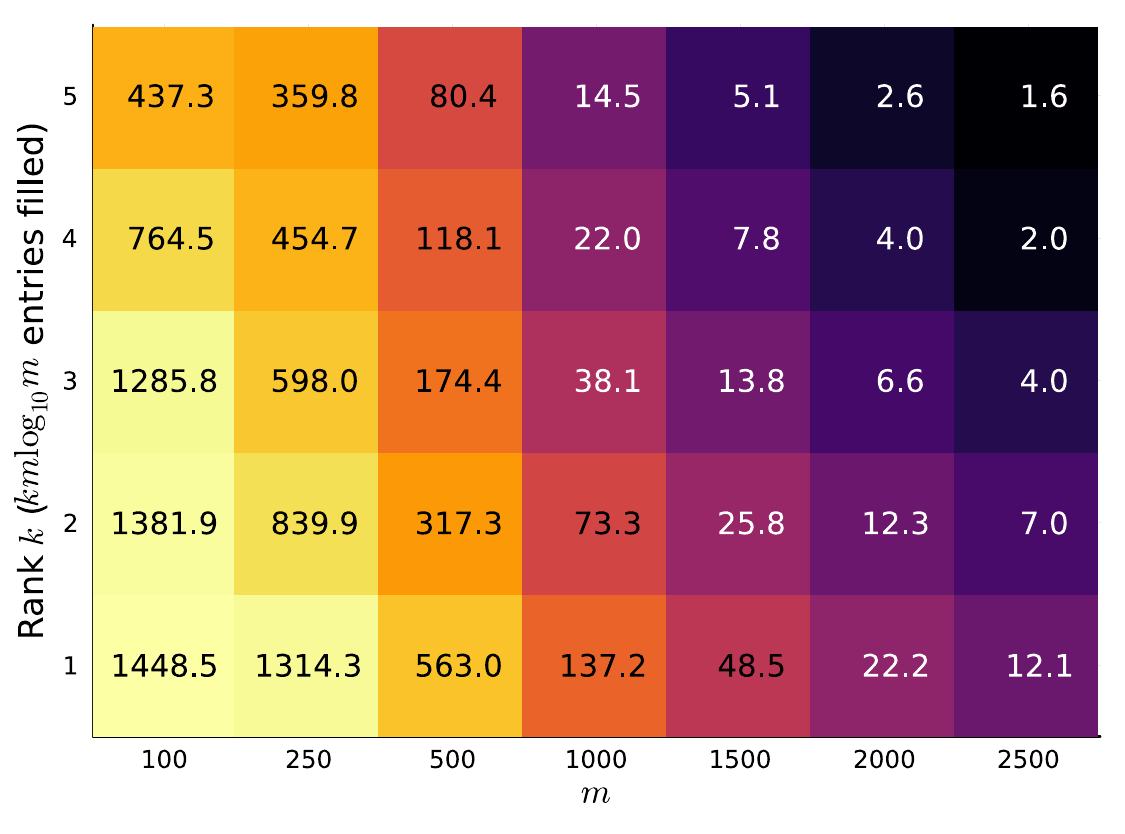}
    \caption{Number of nodes explored}
    \label{fig:mc_scale_nodes_explored_heatmap}
\end{subfigure}
\caption{{\color{black} SDP relaxation solve time in seconds (left) and number of branch-and-bound nodes explored (right) for rank-$k$ matrix completion problems of dimension $50 \times m$, with $km \log_{10}(m)$ filled entries, varying $m$ and $k$, with $\gamma = 120.0$, averaged over 10 random instances.}}
\label{fig:mc_scale_relaxation_time_nodes_explored_heatmap}
\end{figure}

As the rank $k$ increases (and to a lesser extent, as $m$ increases), the difference between the gap at the root node (left panel) and at termination (right panel) is less acute. {\color{black} However, this observation does not necessarily imply that our algorithm is less effective as $k$ and $m$ grow, for at least two reasons.} 
First, the root node gap decreases with $m$ {\color{black} and $k$}, 
providing less {\color{black} opportunity for branch-and-bound to improve}; for example, {\color{black} for $k=2$, the gap at the root node decreases from 38\% when $m = 100$ to 9\% when $m = 2500$.} 
Second, {\color{black} our branch-and-bound algorithm is terminated after a fixed time limit (three hours in all our experiments).} The time needed to solve each SDP relaxation, however, increases with $k$ (and, to a lesser extent, with $m$), {\color{black} as reported on the left panel of Figure~\ref{fig:mc_scale_relaxation_time_nodes_explored_heatmap}. As a result, our branch-and-bound algorithm can explore a smaller number of nodes within the time limit when $k$ or $m$ is larger (see right panel of Figure~\ref{fig:mc_scale_relaxation_time_nodes_explored_heatmap}). Consequently, the observation that the improvement over the root node gap is less acute when $k$ and $m$ increase could also be an artifact of using the same time limit for all instance sizes. This limitation is hard to overcome in practice: stopping the algorithm after a fixed number of nodes explored would suffer from the same limitation, and it is hard to know how to increase time limit (or node limit) with $k$ and $m$ to allow for a `fair' comparison across $(m,k)$ values. For example, it is quite surprising to see that for $k=1,2$ and $m \geq 2000$, exploring 10--20 nodes only can already bring significant gap improvements. 
}

\begin{figure}
\centering
\includegraphics[width=0.6\linewidth]{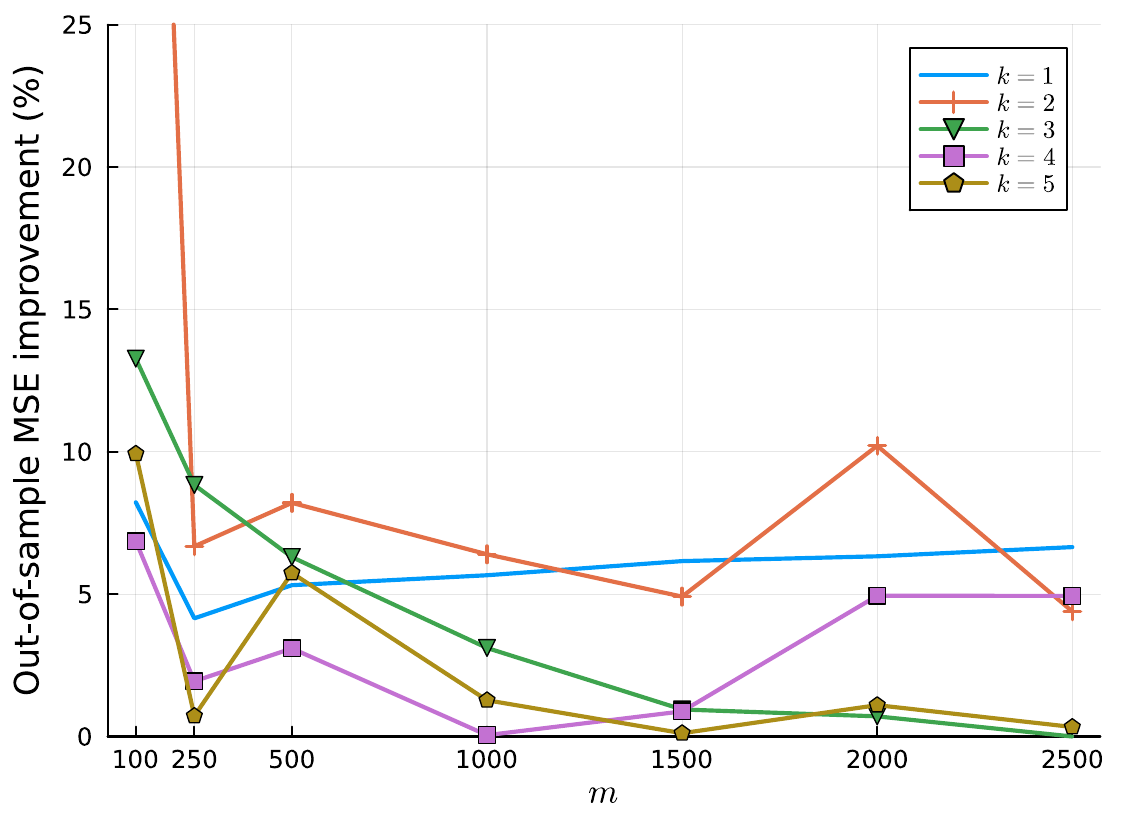}
\caption{Percentage improvement in out-of-sample MSE for rank-$k$ $50 \times m$ matrix completion problems, with $k m \log_{10}(m)$ filled entries, varying $m$ and $k$, with $\gamma = 120.0$, averaged over 10 random instances.}
\label{fig:mc_scale_MSE_out_improvement}
\end{figure}

{\color{black}
We further demonstrate that the solutions found by alternating minimization over branch-and-bound subregions are also superior in terms of downstream out-of-sample error. 
}
Figure~\ref{fig:mc_scale_MSE_out_improvement} (and Table~\ref{tab:mc_scale_MSE_out_improvement}) compares the solution found via alternating minimization at the root node against the best solution found by our branch-and-bound scheme after branching for three hours, as measured by the percentage improvement in out-of-sample mean squared error (Figure~\ref{fig:mc_scale_MSE_out_line_plot} shows the before-and-after values of the out-of-sample MSE). 
We observe that the improved solutions found by branch-and-bound translate to improvements in MSE across the board. 
The improvements are consistently in the 1-10\% range, and the largest improvements are when $m$ is small and Burer-Monteiro at the root node yields poor-quality solutions. 
In short, branch-and-bound improves the test set performance of our imputed matrices. 

\subsection{Summary of Findings from Numerical Experiments} 

The main findings from our numerical experiments are as follows:
\begin{itemize}
    \item Section \ref{ssec:numres.presolve} demonstrates that our valid inequalities for matrix completion problems strengthen, often significantly, the semidefinite relaxation \eqref{eq:mpco_matrix_completion_mprt_relaxed}, and indeed routinely improve the root node gap by an order of magnitude or more in the rank-one case. 
    However, their efficiency depends on the number of observed entries, especially at larger ranks.
    Future work could investigate more {\color{black}scalable semidefinite relaxations that are applicable when $k>1$}.
    \item Section \ref{ssec:bnbdesign} investigates the impact of design choices in our branch-and-bound scheme, and demonstrates that eigenvector-based disjunctions obtain optimality gaps over an order of magnitude smaller than McCormick-based ones in the same amount of computational time.
    {\color{black}
    \item Section \ref{ssec:scalability} demonstrates the scalability of our branch-and-bound scheme for matrix completion problems. While scalability depends on the rank, data regime and extent of regularization, we show that we can solve rectangular $50 \times m$ instances with $m$ as large as $2500$ in hours. One necessary limitation of branch-and-bound is memory; our method requires $\mathcal{O}(K \min\{m, n\}^2)$ memory, where $K$ is the number of nodes explored.
    }
    \item We observe that, at the information-theoretic threshold identified by \citet{candes2009exact} with $O(n k \log n)$ entries observed, our branch-and-bound scheme obtains low-rank matrices with an out-of-sample predictive power {\color{black} $1\%$--$50\%$} better than solutions obtained via a Burer-Monteiro heuristic, depending on the rank and the dimensionality of the problem. 
\end{itemize}

\section{Conclusion}
In this paper, we propose a new branch-and-bound scheme for solving low-rank matrix completion problems to certifiable optimality. The framework considers matrix perspective relaxations and recursively partitions their feasible regions using eigenvector disjunctions. We provide theoretical and empirical evidence to justify the superiority of eigenvector disjunctions compared with the widely used McCormick disjunctions. We further strengthen the semidefinite relaxations via valid inequalities. On rank-one matrix completion problems, we numerically find that these inequalities routinely improve the root node gap by an order of magnitude. 

Altogether,  our approach scales to matrix completion problems on $n \times m$ matrices with $\max\{m, n\} = 2500$ and rank up to $5$ with optimality guarantees.
Compared with the existing and popular scalable heuristics (which, unfortunately, do not provide any {\color{black}instance-wise optimality certificates}), our branch-and-bound scheme obtains low-rank matrices with an out-of-sample reconstruction error $1\%$--$50\%$ lower, depending on the rank and the dimensionality of the problem, illustrating the practical value of certifiably optimal methods for solving matrix completion problems.

\subsection*{Acknowledgments}
Ryan Cory-Wright gratefully acknowledges the MIT-IBM Research Lab for hosting him while part of this work was conducted. 



{\scriptsize
\bibliographystyle{informs2014}

}
\newpage

\ECSwitch
\ECHead{Electronic Companion}

{\color{black}
\section{Regularization for Matrix Completion}
\label{ssec:regularization}
In this section, we provide additional motivation for the presence of the Frobenius regularization $\frac{1}{2\gamma} \| \bm{X} \|^2_F$ in the objective of our matrix completion problem.

In Problem \eqref{eq:rco_matrix_completion}, we introduce an additional penalty on the squared Frobenius norm of $\bm{X}$, in line with previous work on regularized matrix completion. For example, several matrix completion formulations introduce a low-rank decomposition of $\bm{X}$, $\bm{X} = \bm{UV}$ with $\bm{U} \in \mathbb{R}^{n \times k}, \bm{V} \in \mathbb{R}^{k \times m}$ together with a penalty on the squared Frobenius norm of the factors $\bm{U}$ and $\bm{V}$ \citep[see, e.g.,][]{srebro2004maximum,mnih2007probabilistic,zhou2008large}. This regularization on the factors $\bm{U}$ and $\bm{V}$ can be related to a nuclear norm regularization on $\bm{X}$ \citep[][Lemma 1]{srebro2004maximum}---note that the nuclear norm is often used as a convex surrogate for rank \citep[see, e.g.,][]{candes2009exact,candes2010matrix,mazumder2020matrix} but, in these works, this regularization term is introduced in addition to an explicit low-rank decomposition.
Motivated by the ElasticNet regularization for sparse linear regression, \citet{sun2012calibrated} consider a matrix completion problem with a penalty on the nuclear norm and on the squared Frobenius norm of $\bm{X}$. 
Similarly, and closer to our formulation, \citet{keshavan2010regularization} consider a matrix completion problem with an explicit rank requirement and a squared Frobenius penalty on $\bm{X}$. 

Furthermore, we can connect the optimal objective value of the regularized and the unregularized matrix completion problems:  
\begin{proposition}\label{prop.gammaineq}
Let $v(\gamma)$ denote the optimal objective value of Problem~\eqref{eq:rco_matrix_completion}, with minimizer $\bm{X}^\star(\gamma)$ and let $v^\star$ denote the optimal objective value of the unregularized problem with minimizer $\bm{X}^\star$. We have the following inequality
\begin{align}
    v(\gamma) - \frac{1}{2\gamma} \| \bm{X}^\star \|^2_F \leq v^\star \leq v(\gamma)
\end{align}
\end{proposition}

Hence, for a fixed input data $\bm{A}$, we have $v(\gamma) \to v^\star$ as $\gamma \to \infty$. This shows that the regularized objective can approximate the nominal objective to arbitrary precision, by taking $\gamma \to \infty$.
\proof{Proof of Proposition \ref{prop.gammaineq}} To show the first inequality, we use the fact that $\bm{X}^\star$ is feasible for the minimization problem~\eqref{eq:rco_matrix_completion} to write
\begin{align*}
    v(\gamma) \leq \frac{1}{2 \gamma} \| \bm{X}^\star \|^2_F  
    + \frac{1}{2} \sum_{(i,j) \in \mathcal{I}} (X^\star_{i,j} - A_{i,j})^2
    = \frac{1}{2 \gamma} \| \bm{X}^\star \|^2_F  + v^\star.
\end{align*}
For the second inequality, we use the fact that $\bm{X}^\star(\gamma)$ is feasible for the unregularized problem and write
\begin{align*}
    v^\star \leq  \frac{1}{2} \sum_{(i,j) \in \mathcal{I}} (X^\star(\gamma)_{i,j} - A_{i,j})^2 \leq \frac{1}{2 \gamma} \| \bm{X}^\star(\gamma) \|^2_F + \frac{1}{2} \sum_{(i,j) \in \mathcal{I}} (X^\star(\gamma)_{i,j} - A_{i,j})^2 = v(\gamma). \Halmos
\end{align*} 
\endproof

}
\section{Proof of Proposition \ref{prop:mccormickisbad_k}}\label{ssec:proofmccormick}

\proof{Proof of Proposition \ref{prop:mccormickisbad_k}}
It suffices to show that every feasible solution to Problem \eqref{eq:mpco_matrix_completion_mprt_relaxed} can be mapped to a feasible solution to \eqref{eq:mpco_matrix_completion_mprt_relaxed_U2} with the same objective value. Therefore, let $(\bm{Y}, \bm{U})$ be a feasible solution to Problem \eqref{eq:mpco_matrix_completion_mprt_relaxed}.
{\color{black} For $(i,j) \in [n] \times [k]$, $U_{i,j}$} satisfies a constraint $U_{i,j} \in [\underline{U}_{i,j}, \overline{U}_{i,j}]$. {\color{black} If $i \in \mathcal{I}(j)$, this corresponds to a branch of the McCormick disjunction (and we omit the index $t$ of that particular branch for concision), while if $i \notin \mathcal{I}(j)$, it is simply $U_{i,j} \in [-1, 1]$ (i.e., $\underline{U}_{i,j}=-1 $ and $\overline{U}_{i,j} =1)$}. 
Then, we need to construct a $\bm{V} \color{black} \in \mathbb{R}^{n \times k \times k}$ {\color{black} such that $(\bm{Y}, \bm{U}, \bm{V})$} satisfies the constraints in \eqref{eq:mpco_matrix_completion_mprt_relaxed_U2} to establish the result. 

{\color{black} Take $j \in [k]$. For any $i \in [n]$, we need $(V_{i,j,j},{U}_{i,j},{U}_{i,j}) \in \mathcal{M}(\underline{U}_{i,j}, \overline{U}_{i,j},\underline{U}_{i,j}, \overline{U}_{i,j})$.} We set $V_{i,j,j}=U_{i,j}^2$ for any $i \in \mathcal{I}(j)$ {\color{black} (which clearly satisfies the McCormick constraint).} 
{\color{black} If $i \notin \mathcal{I}(j)$, the McCormick constraint $(V_{i,j,j},{U}_{i,j},{U}_{i,j}) \in \mathcal{M}(-1,1,-1,1)$ reduces to $V_{i,j,j} \in [2|U_{i,j}|-1,1]$. We distinguish two cases--note that one of these must hold since $\vert \mathcal{I}(j)\vert \leq n-1$ and hence there exists some $i \notin \mathcal{I}(j)$}
{\color{black} 
\paragraph{Case 1: Suppose that $\sum_{i\notin \mathcal{I}(j)} U_{i,j}^2 > 0$.} Then, first, we observe that the constraint $\bm{I}\succeq  \bm{U}\bm{U}^\top$ implies $\sum_{i \in [n]} U_{i,j}^2 \leq 1$. Thus, define $\alpha \geq 1$ such that 
$\sum_{i\in \mathcal{I}(j)} U_{i,j}^2 + \alpha \sum_{i\notin \mathcal{I}(j)} U_{i,j}^2 = 1$.
We set $V_{i,j,j}= \alpha U_{i,j}^2$ for any $i \notin \mathcal{I}(j)$. We have $V_{i,j,j} \geq U_{i,j}^2 \geq  2 |U_{i,j}| -1$ because $\alpha \geq 1$, and $V_{i,j,j} \leq 1$.
\paragraph{Case 2: If $\sum_{i\notin \mathcal{I}(j)} U_{i,j}^2 = 0$} (hence, $U_{i,j} = 0, \forall i\notin \mathcal{I}(j)$), 
denote $\delta = 1 - \sum_{i \in \mathcal{I}(j)} U_{i,j}^2 \in [0,1]$ and set $V_{i,j,j} = \delta / (n - |\mathcal{I}(j)|) \in [0,1] \subseteq [2 |U_{i,j}| -1, 1]$. 
} 

In both cases, we have that {\color{black} $V_{i,j,j} \in  [2 |U_{i,j}| -1, 1]$ as well as} $\sum_{i=1}^n V_{i,j,j}=1$. 


Similarly, {\color{black} consider $j_1,j_2 \in [k]$ with $j_1 \neq j_2$ and} set $V_{i, j_1,j_2}=U_{i,j_1}U_{i,j_2}$ for any $i \in \mathcal{I}(j_1) \cup \mathcal{I}(j_2)$. {\color{black} For these indices $i$, $(V_{i,j_1,j_2},U_{i,j_1},U_{i,j_2}) \in \mathcal{M}(\underline{U}_{i,j_1}, \overline{U}_{i,j_1},\underline{U}_{i,j_2}, \overline{U}_{i,j_2})$ by definition. For $i \notin  \mathcal{I}(j_1) \cup \mathcal{I}(j_2)$---for concision, we will denote $\mathcal{F} := [n] \setminus (\mathcal{I}(j_1) \cup \mathcal{I}(j_2))$---we need to find a value $V_{i,j_1,j_2}$ such that $V_{i,j_1,j_2} \in [|U_{i,j_1} + U_{i,j_2}| - 1, 1 - |U_{i,j_1} - U_{i,j_2}|]$ and $\sum_{i \in [n]} V_{i,j_1,j_2} = \sum_{i \in  \mathcal{I}(j_1) \cup \mathcal{I}(j_2)} U_{i,j_1} U_{i,j_2} + \sum_{i \in \mathcal{F}} V_{i,j_1,j_2} = 0$. Equivalently, we need to show that $-\sum_{i \in  \mathcal{I}(j_1) \cup \mathcal{I}(j_2)} U_{i,j_1} U_{i,j_2} \in [L,U]$ with $L := \sum_{i \in \mathcal{F}}(|U_{i,j_1} + U_{i,j_2}| - 1)$ and $U := \sum_{i \in \mathcal{F}}(1 - |U_{i,j_1} - U_{i,j_2}|)]$. 

Next, let us denote $\bm{U}_{\cdot,j}$ the $j$-th column of $\bm{U}$. 
Since $\bm{UU^\top} \preceq \bm{I}$, we have $\left|\sum_{i \in  \mathcal{I}(j_1) \cup \mathcal{I}(j_2)} U_{i,j_1} U_{i,j_2} \right|\leq \Vert \bm{U}_{\cdot,j_1}\Vert_2 \Vert \bm{U}_{\cdot,j_2}\Vert_2 \leq 1$ which implies that $-\sum_{i \in  \mathcal{I}(j_1) \cup \mathcal{I}(j_2)} U_{i,j_1} U_{i,j_2} \in [-1,1]$. Hence, it is sufficient to show that, under the assumption that $|\mathcal{F}| = n-| \mathcal{I}(j_1) \cup \mathcal{I}(j_2) | \geq 4$, we have $L \leq -1$ and $U \geq 1$. 

We have
\begin{align*}
    L = & \sum_{i \in \mathcal{F}}(|U_{i,j_1} + U_{i,j_2}| - 1) = \sum_{i \in \mathcal{F}}|U_{i,j_1} + U_{i,j_2}| - |\mathcal{F}| \leq  \sqrt{\vert \mathcal{F}\vert}\sqrt{\sum_{i \in \mathcal{F}}(U_{i,j_1} + U_{i,j_2})^2}- |\mathcal{F}|\\
    & \leq \sqrt{2|\mathcal{F}|} - |\mathcal{F}|,
\end{align*}
where the first inequality holds due to Cauchy-Schwarz, and the second inequality holds because $\Vert \bm{U}_{\cdot,j_1}\pm \bm{U}_{\cdot,j_2}\Vert_2^2 \leq 2$. Hence, for $|\mathcal{F}| \geq 4$, we have $L \leq -1$.
Similarly, for $|\mathcal{F}| \geq 4$, we have
\begin{align*}
    \sum_{i \in \mathcal{F}}(1 - |U_{i,j_1} - U_{i,j_2}|) &\geq |\mathcal{F}| - \sqrt{2|\mathcal{F}|} \geq 1.
\end{align*}
}
Therefore, $(\bm{Y}, \bm{U}, \bm{V})$ is feasible within one branch of our disjunction, and there is thus no hope of improving upon the root node relaxation {\color{black} with this disjunction.} 
\hfill\Halmos
\endproof

{\color{black}
\section{Branch-and-bound design decisions}
\label{appendix:BnB-decisions}

In this section, we provide more detail on the design decisions we make in proposing our branch-and-bound algorithm, namely node selection (\ref{appendix:node_selection}) and branching strategy (\ref{appendix:branching_strategy}).
}

\subsection{Node selection: depth-first vs. breadth-first vs. best-first search}
\label{appendix:node_selection}

One of the most significant design decisions in a branch-and-bound scheme is the node selection strategy employed. 
The three node selection heuristics that we consider in this work are depth-first search (where nodes are selected in a last-in-first-out manner), breadth-first search (where nodes are selected in a first-in-first-out manner), and best-first search (where the node with the lowest remaining lower bound is selected at each iteration). 

Breadth-first and best-first search ensure that the overall lower bound increases at most iterations, provided no ties exist. 
As argued by \citet{lawler1966branch}, best-first search is potentially advantageous, because if the set of branching directions is fixed, then any nodes expanded under this strategy must also be expanded under any other strategy. 
However, both strategies incur a high memory cost from maintaining many unexplored nodes in the queue. 
On the other hand, depth-first search maintains a queue size which is a linear function of the problem size throughout the entire search process \citep{ibaraki1976theoretical}, although at the price of spending less time tightening the upper bound and, therefore, often needed to expand more nodes overall. 
We also observe empirically in Section~\ref{sec:numres} that best-first search usually performs better compared to the other strategies; therefore, we propose using best-first search unless the instance generates too many nodes to fit in available memory, at which point the depth-first strategy should be used.

Finally, while we assumed a breadth-first strategy to establish the convergence of Algorithm \ref{alg:mpco_matrix_completion_linear_cuts} in Theorem \ref{thm:convergence}, the convergence result holds with any node selection strategy.

\subsection{Branching strategy}
\label{appendix:branching_strategy}


When running our branch-and-bound algorithm on a rank-$k$ matrix, each node that is not fathomed generates $q^k$, $q \geq 2$, child nodes corresponding to the $q^k$ regions of the disjunction. 
For example, the disjunction \eqref{disjunction2k} generates $2^k$ child nodes, but, as explained for the rank-one case in Section \ref{ssec:disjgeneric}, more fine-grained upper approximations of the $\ell_2^2$ norm lead to disjunctions over $3^k$ or $4^k$ regions. 

Accordingly, another algorithmic design decision is selecting the number of pieces $q\geq 2$ which should be used 
in our disjunctive branching scheme. 
Indeed, increasing the number of child nodes generated at each iteration, $q^k$, improves the tightness of the bound at the expense of additional computational time for solving all $q^k$ subproblems. 
We investigate this tradeoff numerically in Section \ref{sec:numres}.

\section{Numerical results}
\setlength{\tabcolsep}{8pt}

\subsection{Root Node: Strengthened Relaxations}
\label{appendix:numres.presolve}

{\color{black}
Tables~\ref{tab:mc1_root_solve_time}--\ref{tab:mc1_root_root_node_gap} illustrate the trade-off between optimality gap at the root node (the upper bound is obtained via alternating minimization) and computational time for solving the strengthened relaxation in rank-one matrix completion. For each parameter setting, we either impose no Shor LMIs, all Shor LMIs on the minors in $\mathcal{M}_4$ (4 filled entries), all Shor LMIs in $\mathcal{M}_4$ and half of the Shor LMIs in $\mathcal{M}_3$, and all Shor LMIs in $\mathcal{M}_4$ and $\mathcal{M}_3$. From Table~\ref{tab:mc1_root_solve_time}, we see that larger instances (larger $n$) with more observed entries (larger $p$) are computationally more challenging, and that imposing more Shor LMIs can significantly increase the computation time required. 
In return, Table~\ref{tab:mc1_root_root_node_gap} indicates that imposing more Shor LMIs can yield tighter semidefinite relaxations, successfully narrowing the root node relative gap in a majority of trials. This effect holds across all sizes, regularization parameters, and sparsity settings. Whilst the (arithmetic mean of the) root node gap appears to decrease marginally with the inclusion of all minors in $\mathcal{M}_4$ and $\mathcal{M}_3$ as compared to without, these averages are distorted by the large root node gaps of a few instances which see little benefit from the inclusion of $\mathcal{M}_3$ minors. Including constraints on all minors typically increases the fraction of instances with a root node gap lower than  $10^{-4}$ by 10 to 50 percentage points (a 20--83\% improvement), and this effect is stronger as the number of filled entries increases (larger $p$) and regularization decreases (larger $\gamma$).
Including a subset of the minors in $\mathcal{M}_4 \cup \mathcal{M}_3$ provides a finer control of this trade-off between computation time and relaxation strength.
}


\begin{table}\footnotesize
    \centering
    \begin{tabular}{
        S[table-format=3] 
        S[table-format=1.1] S[table-format=2.1] 
        *{4}{S[table-format=3.4]}
    }
        \toprule
        & & 
        & \multicolumn{4}{c}{Time taken (s)}
        \\
        \cmidrule(lr){4-7} 
        {$n$} & {$p$} & {$\gamma$} 
        & {None} 
        & {$\mathcal{M}_4$} 
        & {$\mathcal{M}_4 \cup \mathcal{M}_3$ (half)} 
        & {$\mathcal{M}_4 \cup \mathcal{M}_3$}
        \\
        \midrule
        10 & 2.0 & 20.0 & 0.01168 & 0.02891 & 0.1078 & 0.1553 \\ 
           &     & 80.0 & 0.01234 & 0.02735 & 0.1227 & 0.2063 \\ 
           & 3.0 & 20.0 & 0.01206 & 0.04775 & 0.3827 & 0.7328 \\ 
           &     & 80.0 & 0.01302 & 0.04504 & 0.4023 & 0.7577 \\ 
        \midrule
        20 & 2.0 & 20.0 & 0.07437 & 0.3082 & 1.505 & 2.854 \\ 
           &     & 80.0 & 0.075 & 0.2884 & 1.678 & 4.091 \\ 
           & 3.0 & 20.0 & 0.07291 & 0.7584 & 6.02 & 21.57 \\ 
           &     & 80.0 & 0.0772 & 0.6874 & 7.617 & 21.73 \\ 
        \midrule
        30 & 2.0 & 20.0 & 0.329 & 2.161 & 8.45 & 19 \\ 
           &     & 80.0 & 0.3629 & 1.637 & 10.11 & 22.88 \\ 
           & 3.0 & 20.0 & 0.3502 & 3.234 & 47.37 & 177.3 \\ 
           &     & 80.0 & 0.3723 & 4.21 & 56.22 & 177.9 \\ 
        \midrule
        50 & 2.0 & 20.0 & 3.066 & 29.24 & 94.18 & 216.5 \\ 
           &     & 80.0 & 3.357 & 25.38 & 119.4 & 271.4 \\ 
           & 3.0 & 20.0 & 3.244 & 30.12 & 884 & 2787 \\ 
           &     & 80.0 & 3.515 & 46.43 & 1090 & 2628 \\ 
        \midrule
        75 & 2.0 & 20.0 & 22.85 & 238.7 & 1002 & 2073 \\ 
           &     & 80.0 & 24.62 & 253.7 & 1070 & 2473 \\ 
           & 3.0 & 20.0 & 25.36 & 266.7 & {--} & {--} \\ 
           &     & 80.0 & 27.07 & 395.5 & {--} & {--} \\ 
        \midrule
        100 & 2.0 & 20.0 & 111.2 & 1628 & {--} & {--} \\ 
            &     & 80.0 & 121.7 & 1386 & {--} & {--} \\ 
            & 3.0 & 20.0 & 112.1 & 1298 & {--} & {--} \\ 
            &     & 80.0 & 125.4 & 1931 & {--} & {--} \\ 
        \bottomrule
        \\
    \end{tabular}%
    \caption{\color{black} Time taken to solve the root node relaxation across rank-one matrix completion problems with $pn \log_{10}(n)$ filled entries, with different Shor LMIs added (arithmetic mean taken over 20 instances per row). A ``--'' indicates parameter settings that did not complete within the 1-hour time limit.}
    \label{tab:mc1_root_solve_time}
\end{table}

\begin{table}\footnotesize
    \centering
    \begin{tabular}{
        S[table-format=3] 
        S[table-format=1.1] S[table-format=2.1] 
        S[table-format=1.2e-1, retain-zero-exponent=true] @{\hspace{4pt}} r
        S[table-format=1.2e-1, retain-zero-exponent=true] @{\hspace{4pt}} r
        S[table-format=1.2e-1, retain-zero-exponent=true] @{\hspace{4pt}} r
        S[table-format=1.2e-1, retain-zero-exponent=true] @{\hspace{4pt}} r
    }
        \toprule
        & & 
        & \multicolumn{8}{c}{Root node gap} 
        \\
        \cmidrule(lr){4-5}
        \cmidrule(lr){6-7}
        \cmidrule(lr){8-9}
        \cmidrule(lr){10-11}
        & & 
        & \multicolumn{2}{c}{None} 
        & \multicolumn{2}{c}{$\mathcal{M}_4$} 
        & \multicolumn{2}{c}{$\mathcal{M}_4 \cup \mathcal{M}_3$ (half)} 
        & \multicolumn{2}{c}{$\mathcal{M}_4 \cup \mathcal{M}_3$}
        \\
        \midrule
        10 & 2.0 & 20.0 & 3.70e-01 &  (5\%) & 3.68e-01 &  (5\%) & 3.64e-01 & (25\%) & 3.56e-01 & (25\%) \\ 
           &     & 80.0 & 1.61e+00 &  (0\%) & 1.59e+00 &  (0\%) & 1.54e+00 &  (5\%) & 1.15e+00 & (10\%) \\ 
           & 3.0 & 20.0 & 1.73e-01 & (10\%) & 1.68e-01 & (15\%) & 1.51e-01 & (40\%) & 1.47e-01 & (50\%) \\ 
           &     & 80.0 & 7.11e-01 &  (0\%) & 6.51e-01 &  (0\%) & 5.61e-01 & (20\%) & 5.40e-01 & (45\%) \\ 
        \midrule
        20 & 2.0 & 20.0 & 1.14e-01 & (15\%) & 1.36e-01 & (25\%) & 1.10e-01 & (45\%) & 1.31e-01 & (55\%) \\ 
           &     & 80.0 & 1.65e-01 &  (0\%) & 1.53e-01 &  (0\%) & 1.23e-01 &  (0\%) & 1.07e-01 & (15\%) \\ 
           & 3.0 & 20.0 & 6.17e-02 & (10\%) & 5.81e-02 & (25\%) & 5.48e-02 & (65\%) & 5.46e-02 & (70\%) \\ 
           &     & 80.0 & 2.71e-01 &  (0\%) & 2.23e-01 &  (0\%) & 1.70e-01 & (60\%) & 1.69e-01 & (70\%) \\ 
        \midrule
        30 & 2.0 & 20.0 & 3.93e-02 & (15\%) & 3.90e-02 & (15\%) & 3.63e-02 & (45\%) & 3.60e-02 & (45\%) \\ 
           &     & 80.0 & 1.84e-01 &  (0\%) & 1.79e-01 &  (0\%) & 1.50e-01 &  (0\%) & 1.40e-01 & (30\%) \\ 
           & 3.0 & 20.0 & 2.07e-02 & (25\%) & 1.96e-02 & (40\%) & 1.89e-02 & (70\%) & 1.89e-02 & (70\%) \\ 
           &     & 80.0 & 1.17e-01 &  (0\%) & 9.68e-02 &  (0\%) & 7.06e-02 & (60\%) & 6.99e-02 & (70\%) \\ 
        \midrule
        50 & 2.0 & 20.0 & 6.47e-03 & (45\%) & 6.33e-03 & (45\%) & 5.15e-03 & (60\%) & 4.85e-03 & (70\%) \\ 
           &     & 80.0 & 4.85e-02 &  (0\%) & 4.60e-02 &  (0\%) & 3.02e-02 & (20\%) & 2.62e-02 & (35\%) \\ 
           & 3.0 & 20.0 & 9.72e-03 & (55\%) & 9.40e-03 & (55\%) & 9.01e-03 & (75\%) & 8.78e-03 & (80\%) \\ 
           &     & 80.0 & 5.29e-02 &  (0\%) & 4.41e-02 &  (0\%) & 3.32e-02 & (65\%) & 2.67e-02 & (83\%) \\ 
        \midrule
        75 & 2.0 & 20.0 & 4.26e-04 & (65\%) & 4.07e-04 & (65\%) & 3.94e-04 & (75\%) & 3.94e-04 & (75\%) \\ 
           &     & 80.0 & 1.15e-02 &  (0\%) & 1.01e-02 &  (0\%) & 2.77e-03 & (45\%) & 1.89e-03 & (50\%) \\ 
           & 3.0 & 20.0 & 2.19e-03 & (65\%) & 2.19e-03 & (70\%) & {--} & {--} & {--} & {--} \\ 
           &     & 80.0 & 1.54e-02 &  (0\%) & 1.15e-02 &  (0\%) & {--} & {--} & {--} & {--} \\ 
        \midrule
        100 & 2.0 & 20.0 & 3.22e-03 & (75\%) & 3.21e-03 & (75\%) & {--} & {--} & {--} & {--} \\ 
            &     & 80.0 & 1.75e-02 &  (0\%) & 1.68e-02 &  (0\%) & {--} & {--} & {--} & {--} \\ 
            & 3.0 & 20.0 & 5.52e-03 & (70\%) & 5.52e-03 & (70\%) & {--} & {--} & {--} & {--} \\ 
            &     & 80.0 & 2.21e-02 & (10\%) & 2.05e-02 & (25\%) & {--} & {--} & {--} & {--} \\ 
        \bottomrule
        \\
    \end{tabular}%
    \caption{\color{black} Relative gap at root node across rank-one matrix completion problems with $pn \log_{10}(n)$ filled entries, with different Shor LMIs added (arithmetic mean taken over 20 instances per row). Values in parentheses denote proportion of trials which converged (root node gap less than $10^{-4}$), out of those completed within the time limit. A ``--'' indicates parameter settings which did not complete within the 1-hour time limit.}
    \label{tab:mc1_root_root_node_gap}
\end{table}

\FloatBarrier
\subsection{Branch-and-bound Design Decisions}

In this section, we document the performance of our branch-and-bound scheme with various parameter settings: comparing our {\color{black}eigenvector} disjunctions to a naive McCormick-based approach, changing the order of the nodes explored, and including alternating minimization at child nodes of the search tree. We evaluate all combinations of parameter settings, and record their relative optimality gap and time taken (capped at one hour) for rank-one matrix completion problems with regularization $\gamma \in \{20.0, 80.0\}$ and $p n \log_{10}(n)$ filled entries with $p \in \{2.0, 3.0\}$. The results are shown in Tables \ref{tab:mc1_gap}--\ref{tab:mc1_time}, \ref{tab:mc1_gap_less_lessreg}--\ref{tab:mc1_time_less_lessreg} (less entries $p = 2.0$ and less regularization $\gamma = 80.0$), \ref{tab:mc1_gap_more_morereg}--\ref{tab:mc1_time_more_morereg} (more entries $p = 3.0$ and more regularization $\gamma = 20.0$), and \ref{tab:mc1_gap_more_lessreg}--\ref{tab:mc1_time_more_lessreg} (more entries $p = 3.0$ and less regularization $\gamma = 80.0$) respectively.

The tables show that eigenvector disjunctions perform consistently better than McCormick disjunctions, and that best-first search on unexplored child nodes is usually a good node selection strategy. They also illustrate the power of performing alternating minimization at (some) child nodes, because better feasible solutions can be found with different initializations, which yield tight upper bounds and thereby accelerating the branch-and-bound procedure.

{

\begin{table}\footnotesize
    \centering
    \begin{tabular}{
        S[table-format=2.0] c
        S[table-format=1.2e-1, retain-zero-exponent=true] 
        S[table-format=1.2e-1, retain-zero-exponent=true] 
        S[table-format=1.2e-1, retain-zero-exponent=true]
        S[table-format=1.2e-1, retain-zero-exponent=true] 
        S[table-format=1.2e-1, retain-zero-exponent=true] 
        S[table-format=1.2e-1, retain-zero-exponent=true]
    }
        \toprule 
        & & \multicolumn{3}{c}{With McCormick disjunctions} & \multicolumn{3}{c}{With eigenvector disjunctions} 
        \\
        \cmidrule(r){3-5} \cmidrule(l){6-8}
        {\multirow{2}{*}{$n$}} & {Alternating} 
        & {\multirow{2}{*}{Best-first}} & {\multirow{2}{*}{Breadth-first}} & {\multirow{2}{*}{Depth-first}}
        & {\multirow{2}{*}{Best-first}} & {\multirow{2}{*}{Breadth-first}} & {\multirow{2}{*}{Depth-first}}
        \\
        & {minimization} & & &
        \\
        \midrule
        10 & \xmark & 1.42e+00 & 1.50e+00 & 1.50e+00 & 2.21e+00 & 2.40e+00 & 1.91e+00 \\ 
        10 & \cmark & 7.97e-02 & 1.14e-01 & 6.26e-01 & \cellcolor[gray]{0.9}5.39e-02 & 8.89e-02 & 3.18e-01 \\ 
        \midrule
        20 & \xmark & 3.38e-01 & 3.38e-01 & 3.38e-01 & 5.02e-01 & 5.27e-01 & 4.09e-01 \\ 
        20 & \cmark & 8.61e-02 & 8.77e-02 & 8.77e-02 & \cellcolor[gray]{0.9}5.11e-02 & 6.50e-02 & 2.25e-01 \\ 
        \midrule
        30 & \xmark & 2.34e-01 & 2.34e-01 & 2.34e-01 & 2.34e-01 & 2.42e-01 & 2.18e-01 \\ 
        30 & \cmark & 1.08e-01 & 1.13e-01 & 1.13e-01 & \cellcolor[gray]{0.9}4.71e-02 & 5.42e-02 & 1.04e-01 \\ 
        \midrule
        40 & \xmark & 2.12e-01 & 2.12e-01 & 2.12e-01 & 2.07e-01 & 2.11e-01 & 2.13e-01 \\ 
        40 & \cmark & 1.05e-01 & 1.05e-01 & 1.05e-01 & \cellcolor[gray]{0.9}3.08e-02 & 3.38e-02 & 8.51e-02 \\ 
        \midrule
        50 & \xmark & 7.45e-02 & 7.45e-02 & 7.45e-02 & 1.35e-01 & 1.37e-01 & 1.39e-01 \\ 
        50 & \cmark & 4.53e-02 & 4.53e-02 & 4.53e-02 & \cellcolor[gray]{0.9}2.15e-02 & 2.32e-02 & 6.64e-02 \\ 
        \bottomrule
    \end{tabular}
    \vspace{1pt}
    \caption{Final optimality gap 
    across rank-one matrix completion problems with $|\mathcal{I}| = pn \log_{10}(n)$ filled entries, averaged over 20 instances per row ($p = 2.0$, $\gamma = 80.0$).}
    \label{tab:mc1_gap_less_lessreg}
\end{table}
}

{

\begin{table}\footnotesize
    \centering
    \begin{tabular}{
        S[table-format=2.0] c
        S[table-format=4.1] 
        S[table-format=4.1] 
        S[table-format=4.1]
        S[table-format=4.1] 
        S[table-format=4.1] 
        S[table-format=4.1]
    }
        \toprule 
        & & \multicolumn{3}{c}{With McCormick disjunctions} & \multicolumn{3}{c}{With eigenvector disjunctions} 
        \\
        \cmidrule(r){3-5} \cmidrule(l){6-8}
        {\multirow{2}{*}{$n$}} & {Alternating} 
        & {\multirow{2}{*}{Best-first}} & {\multirow{2}{*}{Breadth-first}} & {\multirow{2}{*}{Depth-first}}
        & {\multirow{2}{*}{Best-first}} & {\multirow{2}{*}{Breadth-first}} & {\multirow{2}{*}{Depth-first}}
        \\
        & {minimization} & & &
        \\
        \midrule
        10 & \xmark & 3424.9 & 3425.3 & 3600.0 & 3430.1 & 3600.0 & 3600.0 \\ 
        10 & \cmark & 3162.1 & 3204.4 & 3600.0 & 3274.8 & 3600.0 & 3600.0 \\ 
        \midrule
        20 & \xmark & 3600.1 & 3600.1 & 3600.1 & 3600.1 & 3600.1 & 3600.1 \\ 
        20 & \cmark & 3600.1 & 3600.1 & 3600.1 & 3600.1 & 3600.1 & 3600.1 \\ 
        \midrule
        30 & \xmark & 3600.4 & 3600.3 & 3600.3 & 3600.3 & 3600.3 & 3600.4 \\ 
        30 & \cmark & 3600.4 & 3600.3 & 3600.2 & 3600.3 & 3600.3 & 3600.4 \\ 
        \midrule
        40 & \xmark & 3601.0 & 3601.1 & 3600.7 & 3600.8 & 3600.9 & 3600.9 \\ 
        40 & \cmark & 3601.2 & 3601.0 & 3600.8 & 3600.9 & 3601.1 & 3600.8 \\ 
        \midrule
        50 & \xmark & 3602.5 & 3602.1 & 3602.1 & 3602.0 & 3602.1 & 3602.7 \\ 
        50 & \cmark & 3602.4 & 3602.1 & 3602.1 & 3601.9 & 3602.5 & 3602.1 \\
        \bottomrule
    \end{tabular}
    \vspace{1pt}
    \caption{Computational time (s) across rank-one matrix completion problems with $| \mathcal{I} | = pn \log_{10}(n)$ entries, using McCormick disjunctions (top), eigenvector disjunctions (bottom), averaged over 20 instances ($p = 2.0$, $\gamma = 80.0$).}
    \label{tab:mc1_time_less_lessreg}
\end{table}
}

{

\begin{table}\footnotesize
    \centering
    \begin{tabular}{
        S[table-format=2.0] c
        S[table-format=1.2e-1, retain-zero-exponent=true] 
        S[table-format=1.2e-1, retain-zero-exponent=true] 
        S[table-format=1.2e-1, retain-zero-exponent=true]
        S[table-format=1.2e-1, retain-zero-exponent=true] 
        S[table-format=1.2e-1, retain-zero-exponent=true] 
        S[table-format=1.2e-1, retain-zero-exponent=true]
    }
        \toprule 
        & & \multicolumn{3}{c}{With McCormick disjunctions} & \multicolumn{3}{c}{With eigenvector disjunctions} 
        \\
        \cmidrule(r){3-5} \cmidrule(l){6-8}
        {\multirow{2}{*}{$n$}} & {Alternating} 
        & {\multirow{2}{*}{Best-first}} & {\multirow{2}{*}{Breadth-first}} & {\multirow{2}{*}{Depth-first}}
        & {\multirow{2}{*}{Best-first}} & {\multirow{2}{*}{Breadth-first}} & {\multirow{2}{*}{Depth-first}}
        \\
        & {minimization} & & &
        \\
        \midrule
        10 & \xmark & 3.35e-01 & 3.55e-01 & 3.57e-01 & 1.73e-01 & 1.73e-01 & 3.13e-01 \\ 
        10 & \cmark & 3.40e-02 & 4.83e-02 & 1.49e-01 & \cellcolor[gray]{0.9}7.16e-03 & 1.29e-02 & 1.15e-01 \\ 
        \midrule
        20 & \xmark & 4.63e-02 & 4.63e-02 & 4.63e-02 & 1.08e-02 & 1.16e-02 & 1.20e-02 \\ 
        20 & \cmark & 9.58e-03 & 2.54e-02 & 2.54e-02 & \cellcolor[gray]{0.9}1.79e-04 & 9.03e-04 & 1.13e-02 \\ 
        \midrule
        30 & \xmark & 6.18e-02 & 6.18e-02 & 6.18e-02 & 9.70e-03 & 9.57e-03 & 4.65e-02 \\ 
        30 & \cmark & 3.99e-03 & 1.56e-02 & 1.56e-02 & \cellcolor[gray]{0.9}1.32e-04 & 2.22e-04 & 3.02e-03 \\ 
        \midrule
        40 & \xmark & 1.27e-02 & 1.27e-02 & 1.27e-02 & 4.43e-03 & 4.55e-03 & 8.10e-03 \\ 
        40 & \cmark & 9.62e-03 & 9.62e-03 & 9.62e-03 & \cellcolor[gray]{0.9}1.96e-04 & 2.89e-04 & 8.44e-03 \\ 
        \midrule
        50 & \xmark & 9.11e-03 & 9.11e-03 & 9.11e-03 & 1.01e-04 & 1.07e-04 & 2.31e-03 \\ 
        50 & \cmark & 7.13e-03 & 7.13e-03 & 7.13e-03 & \cellcolor[gray]{0.9}1.01e-04 & 1.06e-04 & 6.18e-03 \\ 
        \bottomrule
    \end{tabular}
    \vspace{1pt}
    \caption{Final optimality gap 
    across rank-one matrix completion problems with $|\mathcal{I}| = pn \log_{10}(n)$ filled entries, averaged over 20 instances per row ($p = 3.0$, $\gamma = 20.0$).}
    \label{tab:mc1_gap_more_morereg}
\end{table}
}

{

\begin{table}\footnotesize
    \centering
    \begin{tabular}{
        S[table-format=2.0] c
        S[table-format=4.2] 
        S[table-format=4.2] 
        S[table-format=4.2]
        S[table-format=4.2] 
        S[table-format=4.2] 
        S[table-format=4.2]
    }
        \toprule 
        & & \multicolumn{3}{c}{With McCormick disjunctions} & \multicolumn{3}{c}{With eigenvector disjunctions} 
        \\
        \cmidrule(r){3-5} \cmidrule(l){6-8}
        {\multirow{2}{*}{$n$}} & {Alternating} 
        & {\multirow{2}{*}{Best-first}} & {\multirow{2}{*}{Breadth-first}} & {\multirow{2}{*}{Depth-first}}
        & {\multirow{2}{*}{Best-first}} & {\multirow{2}{*}{Breadth-first}} & {\multirow{2}{*}{Depth-first}}
        \\
        & {minimization} & & &
        \\
        \midrule
        10 & \xmark & 2547.4 & 2553.2 & 3060.0 & 2152.4 & 2118.6 & 3060.1 \\ 
        10 & \cmark & 1871.5 & 1885.8 & 3060.0 & \cellcolor[gray]{0.9}1524.9 & 1990.3 & 2809.7 \\ 
        \midrule
        20 & \xmark & 3240.1 & 3240.1 & 3240.1 & 1746.0 & 2260.7 & 2880.1 \\ 
        20 & \cmark & 2704.7 & 3060.1 & 3060.1 & \cellcolor[gray]{0.9}1743.3 & 2262.4 & 2880.1 \\ 
        \midrule
        30 & \xmark & 3060.4 & 3060.3 & 3060.3 & 1340.8 & 1517.6 & 2880.4 \\ 
        30 & \cmark & 2565.5 & 2700.5 & 2700.4 & \cellcolor[gray]{0.9}854.99 & 1429.6 & 2340.5 \\ 
        \midrule
        40 & \xmark & 1981.3 & 1981.3 & 1981.3 & 976.3 & 1546.2 & 1981.2 \\ 
        40 & \cmark & 1981.4 & 1981.5 & 1981.2 & \cellcolor[gray]{0.9}597.66 & 1141.3 & 1981.4 \\ 
        \midrule
        50 & \xmark & 1983.7 & 1983.8 & 1983.3 & 262.55 & 406.79 & 1100.1 \\ 
        50 & \cmark & 1443.7 & 1443.9 & 1443.7 & \cellcolor[gray]{0.9}253.22 & 381.37 & 959.29 \\ 
        \bottomrule
    \end{tabular}
    \vspace{1pt}
    \caption{Computational time (s) across rank-one matrix completion problems with $| \mathcal{I} | = pn \log_{10}(n)$ entries, using McCormick disjunctions (top), eigenvector disjunctions (bottom), averaged over 20 instances ($p = 3.0$, $\gamma = 20.0$).}
    \label{tab:mc1_time_more_morereg}
\end{table}
}

{
\begin{table}\footnotesize
    \centering
    \begin{tabular}{
        S[table-format=2.0] c
        S[table-format=1.2e-1, retain-zero-exponent=true] 
        S[table-format=1.2e-1, retain-zero-exponent=true] 
        S[table-format=1.2e-1, retain-zero-exponent=true]
        S[table-format=1.2e-1, retain-zero-exponent=true] 
        S[table-format=1.2e-1, retain-zero-exponent=true] 
        S[table-format=1.2e-1, retain-zero-exponent=true]
    }
        \toprule 
        & & \multicolumn{3}{c}{With McCormick disjunctions} & \multicolumn{3}{c}{With eigenvector disjunctions} 
        \\
        \cmidrule(r){3-5} \cmidrule(l){6-8}
        {\multirow{2}{*}{$n$}} & {Alternating} 
        & {\multirow{2}{*}{Best-first}} & {\multirow{2}{*}{Breadth-first}} & {\multirow{2}{*}{Depth-first}}
        & {\multirow{2}{*}{Best-first}} & {\multirow{2}{*}{Breadth-first}} & {\multirow{2}{*}{Depth-first}}
        \\
        & {minimization} & & &
        \\
        \midrule
        10 & \xmark & 5.26e-01 & 5.93e-01 & 5.96e-01 & 4.73e-01 & 5.40e-01 & 5.76e-01 \\ 
        10 & \cmark & 1.03e-01 & 1.48e-01 & 2.76e-01 & \cellcolor[gray]{0.9}7.03e-02 & 1.10e-01 & 2.52e-01 \\ 
        \midrule
        20 & \xmark & 4.63e-02 & 4.63e-02 & 4.63e-02 & 1.08e-02 & 1.16e-02 & 1.20e-02 \\ 
        20 & \cmark & 9.58e-03 & 2.54e-02 & 2.54e-02 & \cellcolor[gray]{0.9}1.79e-04 & 9.03e-04 & 1.13e-02 \\ 
        \midrule
        30 & \xmark & 6.18e-02 & 6.18e-02 & 6.18e-02 & 9.70e-03 & 9.57e-03 & 4.65e-02 \\ 
        30 & \cmark & 3.99e-03 & 1.56e-02 & 1.56e-02 & \cellcolor[gray]{0.9}1.32e-04 & 2.22e-04 & 3.02e-03 \\ 
        \midrule
        40 & \xmark & 1.27e-02 & 1.27e-02 & 1.27e-02 & 4.43e-03 & 4.55e-03 & 8.10e-03 \\ 
        40 & \cmark & 9.62e-03 & 9.62e-03 & 9.62e-03 & \cellcolor[gray]{0.9}1.96e-04 & 2.89e-04 & 8.44e-03 \\ 
        \midrule
        50 & \xmark & 9.11e-03 & 9.11e-03 & 9.11e-03 & 1.01e-04 & 1.07e-04 & 2.31e-03 \\ 
        50 & \cmark & 7.13e-03 & 7.13e-03 & 7.13e-03 & \cellcolor[gray]{0.9}1.01e-04 & 1.06e-04 & 6.18e-03 \\ 
        \bottomrule
    \end{tabular}
    \vspace{1pt}
    \caption{Final optimality gap 
    across rank-one matrix completion problems with $|\mathcal{I}| = pn \log_{10}(n)$ filled entries, averaged over 20 instances per row ($p = 3.0$, $\gamma = 80.0$).}
    \label{tab:mc1_gap_more_lessreg}
\end{table}
}

{

\begin{table}\footnotesize
    \centering
    \begin{tabular}{
        S[table-format=2.0] c
        S[table-format=4.1] 
        S[table-format=4.1] 
        S[table-format=4.1]
        S[table-format=4.1] 
        S[table-format=4.1] 
        S[table-format=4.1]
    }
        \toprule 
        & & \multicolumn{3}{c}{With McCormick disjunctions} & \multicolumn{3}{c}{With eigenvector disjunctions} 
        \\
        \cmidrule(r){3-5} \cmidrule(l){6-8}
        {\multirow{2}{*}{$n$}} & {Alternating} 
        & {\multirow{2}{*}{Best-first}} & {\multirow{2}{*}{Breadth-first}} & {\multirow{2}{*}{Depth-first}}
        & {\multirow{2}{*}{Best-first}} & {\multirow{2}{*}{Breadth-first}} & {\multirow{2}{*}{Depth-first}}
        \\
        & {minimization} & & &
        \\
        \midrule
        10 & \xmark & 3329.9 & 3346.5 & 3600.0 & 3600.0 & 3600.0 & 3600.0 \\ 
        10 & \cmark & \cellcolor[gray]{0.9}3228.0 & 3308.0 & 3600.0 & 3600.0 & 3600.0 & 3600.0 \\ 
        \midrule
        20 & \xmark & 3600.1 & 3600.1 & 3600.1 & 3600.1 & 3600.1 & 3600.1 \\ 
        20 & \cmark & 3600.1 & 3600.1 & 3600.1 & 3600.1 & 3600.1 & 3600.1 \\ 
        \midrule
        30 & \xmark & 3600.3 & 3600.4 & 3600.3 & 3600.4 & 3600.3 & 3600.3 \\ 
        30 & \cmark & 3600.3 & 3600.4 & 3600.3 & 3600.4 & 3600.2 & 3600.4 \\ 
        \midrule
        40 & \xmark & 3601.2 & 3600.9 & 3600.8 & 3600.9 & 3600.8 & 3601.1 \\ 
        40 & \cmark & 3601.0 & 3601.0 & 3600.8 & 3601.1 & 3600.9 & 3601.0 \\ 
        \midrule
        50 & \xmark & 3602.4 & 3602.2 & 3602.2 & 3602.5 & 3602.0 & 3602.4 \\ 
        50 & \cmark & 3602.3 & 3602.9 & 3602.0 & 3602.3 & 3602.1 & 3602.9 \\ 
        \bottomrule
    \end{tabular}
    \vspace{1pt}
    \caption{Computational time (s) across rank-one matrix completion problems with $| \mathcal{I} | = pn \log_{10}(n)$ entries, using McCormick disjunctions (top), eigenvector disjunctions (bottom), averaged over 20 instances ($p = 3.0$, $\gamma = 80.0$).}
    \label{tab:mc1_time_more_lessreg}
\end{table}
}

\begin{figure}
    \centering
    \includegraphics[width=0.7\textwidth]{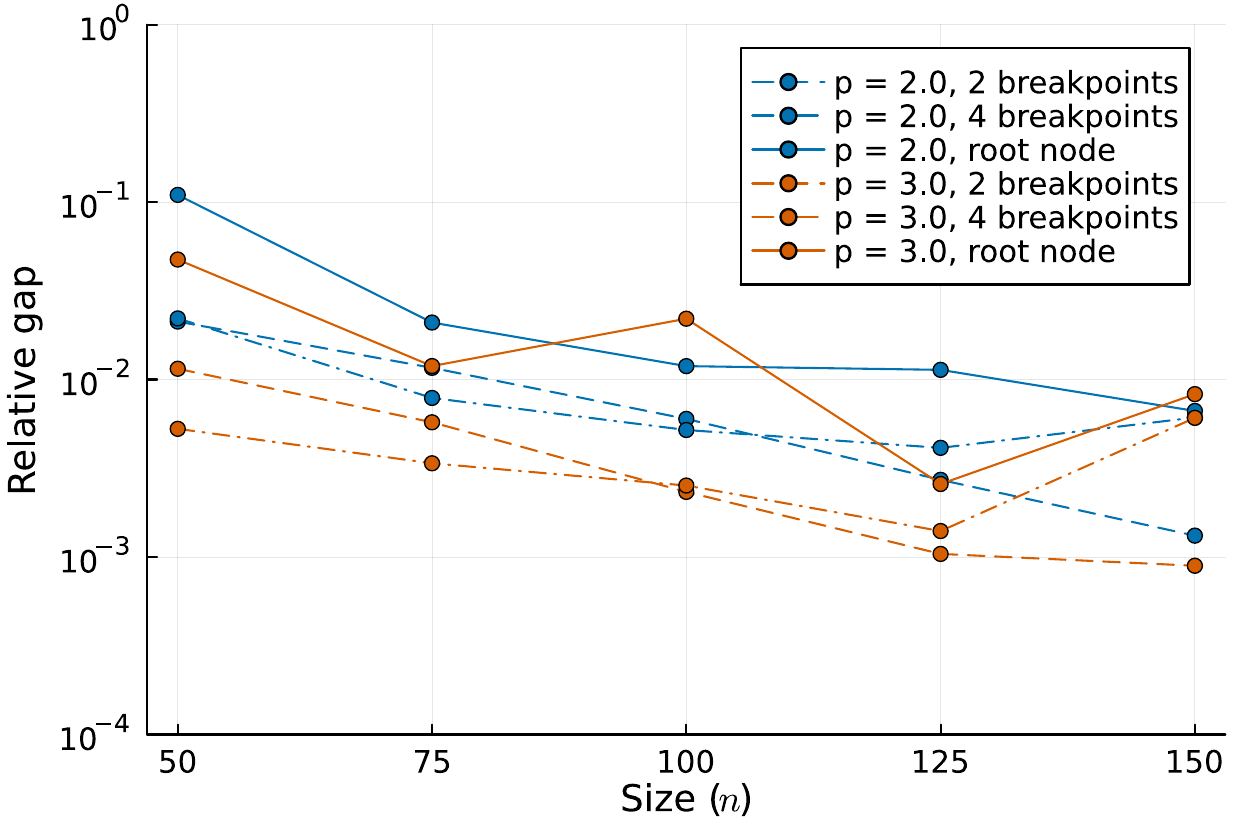}
    \caption{Comparison of relative gap (at root node, and after branching with 2-piece and 4-piece disjunctions) for rank-one matrix completion problems with $pn \log_{10}(n)$ filled entries, after one hour (averaged over 20 instances per point).}
    \label{fig:mc1_size_gap_pknlog_linear_linear3}
\end{figure}

In Figure \ref{fig:mc1_size_gap_pknlog_linear_linear3}, we observe the effect of the number of pieces used $q$ in our disjunctive scheme on the final relative gap after branch-and-bound for one hour on rank-one matrix completion problems with $n \geq 50$. As the problem size $n$ increases, keeping the number of observed entries at $pn \log_{10}(n)$ for $p \in \{2.0, 3.0\}$, the relative advantage of 4-piece disjunctions vanishes -- this is likely due to the fact that, as $n$ increases, the time taken for a single semidefinite relaxation increases and using 4-piece disjunctions introduces more child nodes, leading to an overall increase in computational time. 
Hence, we only recommend using 4-piece disjunctions for rank-one matrix completion problems of low to moderate size.

\FloatBarrier
\subsection{Scalability Experiments}\label{ssec:moreexperiments}

Here we delve into a more detailed investigation of the scalability of our approach for matrix completion problems to large problem sizes and larger ranks. 

\begin{figure}
    \centering
    \includegraphics[width=0.8\linewidth]{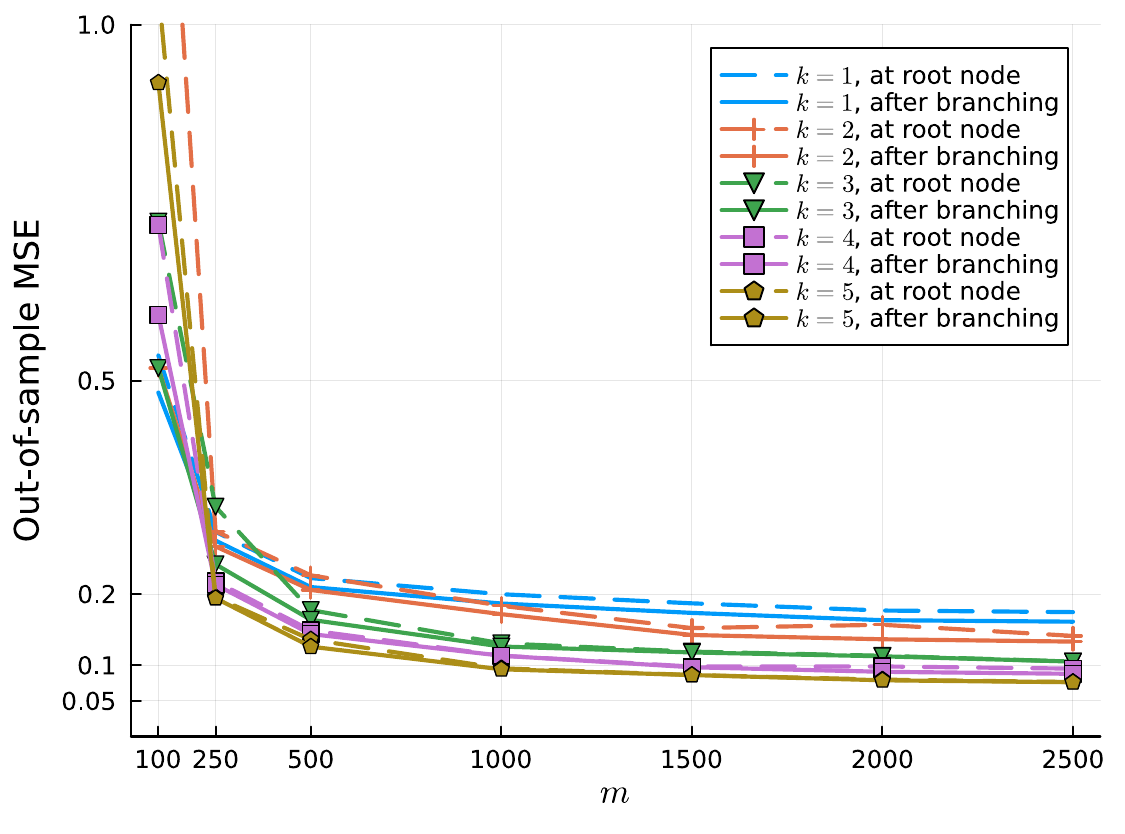}
    \caption{Out-of-sample MSE at the root node and after branching, for rank-$k$ matrix completion problems of dimension $50 \times m$, with $k m \log_{10}(m)$ filled entries, varying $m$ and $k$, with $\gamma = 120.0$, averaged over 10 random instances.}
    \label{fig:mc_scale_MSE_out_line_plot}
\end{figure}

Figure \ref{fig:mc_scale_MSE_out_line_plot} shows the out-of-sample mean-squared error (MSE), at the root node and after branching for three hours, on the large rectangular $50 \times m$ instances. These plots show that branching improves the out-of-sample MSE for all values of $m$, with the most significant improvement in out-of-sample MSE when $m$ is small. 

\begin{table}\footnotesize
    \centering
    \begin{tabular}{
        S[table-format=1]
        S[table-format=1.2e-1]
        S[table-format=1.2e-1]
        S[table-format=1.2e-1]
        S[table-format=1.2e-1]
        S[table-format=1.2e-1]
        S[table-format=1.2e-1]
        S[table-format=1.2e-1]
    }
        \toprule
        & \multicolumn{7}{c}{Relative improvement}
        \\
        \cmidrule(lr){2-8}
        {$k$} 
        & {$m$ = 100} & {$m$ = 250} & {$m$ = 500} & {$m$ = 1000} & {$m$ = 1500} & {$m$ = 2000} & {$m$ = 2500} 
        \\
        \midrule
        1 & 8.22e-02 & 4.15e-02 & 5.31e-02 & 5.66e-02 & 6.16e-02 & 6.33e-02 & 6.65e-02 \\ 
        2 & 5.84e-01 & 6.68e-02 & 8.21e-02 & 6.40e-02 & 4.91e-02 & 1.02e-01 & 4.39e-02 \\ 
        3 & 1.33e-01 & 8.83e-02 & 6.30e-02 & 3.11e-02 & 9.48e-03 & 7.07e-03 & -9.29e-06 \\ 
        4 & 6.86e-02 & 1.94e-02 & 3.09e-02 & 4.71e-04 & 8.82e-03 & 4.94e-02 & 4.94e-02 \\ 
        5 & 9.93e-02 & 7.21e-03 & 5.74e-02 & 1.27e-02 & 1.16e-03 & 1.10e-02 & 3.34e-03 \\ 
        \bottomrule
    \end{tabular}
    \vspace{1pt}
    \caption{Relative improvement in out-of-sample MSE after branching for 3 hours, for rank-$k$ matrix completion problems of dimension $50 \times m$, with $k m \log_{10}(m)$ entries, varying $m$ and $k$, with $\gamma = 120.0$, averaged over 10 random instances.}
    \label{tab:mc_scale_MSE_out_improvement}
\end{table}

\end{document}